\documentclass[letterpaper, 10 pt, conference]{ieeeconf}  

\IEEEoverridecommandlockouts                              
\makeatletter
\let\NAT@parse\undefined
\makeatother

\usepackage[dvipsnames]{xcolor}

\newcommand*\linkcolours{ForestGreen}

\usepackage{times}
\usepackage{graphicx}
\usepackage{amssymb}
\usepackage{gensymb}
\usepackage{amsmath}
\usepackage{breakurl}
\usepackage{caption}
\usepackage{subcaption}
\usepackage{cuted}

\usepackage[utf8]{inputenc}

\usepackage[ruled, linesnumbered]{algorithm2e} 
\usepackage[figuresright]{rotating}
\usepackage{booktabs} 
\usepackage{multirow} 

\usepackage{url,hyperref}
\hypersetup{
colorlinks,
linkcolor=\linkcolours,
citecolor=\linkcolours,
filecolor=\linkcolours,
urlcolor=\linkcolours}

\usepackage[labelfont={bf},font=small]{caption}
\usepackage[none]{hyphenat}

\usepackage{mathtools, cuted}

\usepackage[noadjust, nobreak]{cite}
\usepackage{tabularx}
\usepackage{amsmath}
\usepackage{float}
\usepackage{pifont}

\newcolumntype{Y}{>{\centering\arraybackslash}X}
\usepackage[]{placeins}

\usepackage{placeins}

\usepackage{tikz}

\usepackage[framemethod=tikz]{mdframed}
\usepackage{afterpage}
\usepackage{stfloats}
\usepackage{atbegshi}
\usepackage{amsmath}
\usepackage{graphicx}
\usepackage{subcaption}
\usepackage{multirow}
\newcommand{\handlethispage}{}
\newcommand{\discardpagesfromhere}{\let\handlethispage\AtBeginShipoutDiscard}
\newcommand{\keeppagesfromhere}{\let\handlethispage\relax}
\AtBeginShipout{\handlethispage}

\usepackage{comment}
\title{\LARGE \bf
OneEncoder: A Lightweight Framework for Progressive Alignment 
of Modalities}

\author{Bilal FAYE$^{1}$, Hanane AZZAG$^{2}$, Mustapha Lebbah$^{3}$  \\
	\normalsize e-mail: faye@lipn.univ-paris13.fr, azzag@univ-paris13.fr, mustapha.lebbah@uvsq.fr
}

\begin{document}

\maketitle
\thispagestyle{empty}
\pagestyle{empty}

\begin{abstract}
Cross-modal alignment Learning integrates information from different modalities like text, image, audio and video to create unified models. This approach develops shared representations and learns correlations between modalities, enabling applications such as visual question answering and audiovisual content analysis.
Current techniques rely on large modality-specific encoders, necessitating fine-tuning or training from scratch on vast aligned datasets (e.g., text-image, text-audio, image-audio). This approach has limitations: (i) it is very expensive due to the need for training large encoders on extensive datasets, (ii) acquiring aligned large paired datasets is challenging, and (iii) adding new modalities requires retraining the entire framework to incorporate these modalities.
To address these issues, we propose OneEncoder, a lightweight framework that progressively represents and aligns \textit{\textbf{four}} modalities (image, text, audio, video). Initially, we train a lightweight Universal Projection module (UP) to align image and text modalities. Then, we freeze the pretrained UP and progressively align future modalities to those already aligned. OneEncoder operates efficiently and cost-effectively, even in scenarios where vast aligned datasets are unavailable, due to its lightweight design. Trained on small paired datasets, it shows strong performance in tasks like classification, querying, and visual question answering, surpassing methods that rely on large datasets and specialized encoders.
\end{abstract}


\section{Introduction}
\indent Multimodal processing integrates information from various sensory modalities within a single system. Cross-modal alignment learning (CM-AL) involves developing shared representations to enable seamless understanding across different modalities~\cite{baltru2020,chen2018,srivastava2012}. Recent advancements in Large Language Models~\cite{devlin2019bert,openai2024gpt4}, Large Vision Models~\cite{Dosovitskiy2020,liu2021swin}, and Large Audio Models~\cite{li2019jasper,baevski2020wav2vec}
have enabled CM-AL tasks,
 starting with the alignment of 
two modalities and extending to multiple modalities~\cite{radford2021learning,ramesh2021zero,guzhov2021audioclip,akbari2021vatt}. However, these methods depend on large modality-specific encoders that must be trained on vast aligned datasets to achieve high performance. This approach presents several challenges: large models incur high computational costs; acquiring large aligned datasets, especially for specific modalities, is difficult; furthermore, once trained, adding new modalities requires retraining the entire framework, resulting in significant costs. Thus, a critical challenge is \textit{to develop a framework that achieves good alignment results without relying on vast aligned datasets, while progressively aligning diverse modalities}.\newline
\begin{figure}[htbp]
      \centering
      \includegraphics[width=0.5\textwidth]{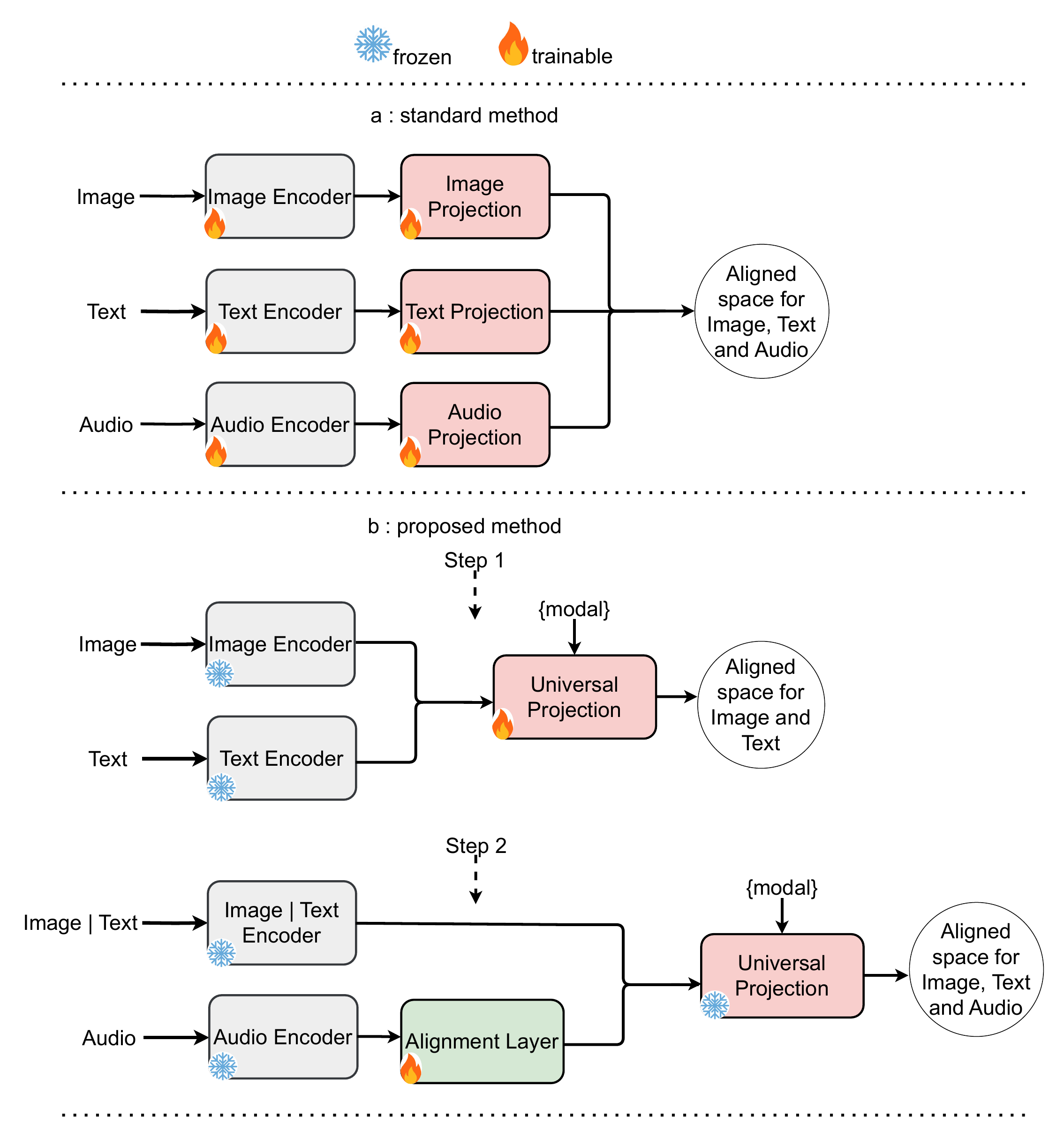}
      \caption{Comparison of three modality alignment methods: Standard cross-modal vs. OneEncoder. Standard aligns via simultaneous training of modality-specific encoders. OneEncoder uses frozen, pretrained encoders with a lightweight Universal Projection (UP) module trained on two modalities. For new modalities, UP stays frozen, training only the Alignment Layer. Modality tokens enable efficient switching between modalities. Using this method, video can be aligned with other modalities (image, text, audio) in the same way.}
      \label{fig:comparison}
\end{figure}
\indent We draw inspiration from recent research on aligning modalities with language by Han et al.~\cite{han2023onellm}. In their OneLLM study, the authors employ a frozen pretrained vision-language encoder to encode eight modalities and align them simultaneously with the text modality using a learned mixture of projection experts. This insight suggests that training modality-specific encoders may not be necessary; instead, pretrained encoders can be utilized to represent specific modalities.
\newline
\indent In this paper, we introduce OneEncoder, which progressively aligns five modalities (image, text, audio and video) within a single unified framework. Illustrated in Figure~\ref{fig:comparison}, OneEncoder comprises frozen pretrained modality-specific encoders, a lightweight Universal Projection module (UP), a compact Alignment Layer (AL),  
and modality tokens to facilitate modality switching~\cite{zhu2023minigpt4,liu2023visual,li2023blip,han2023onellm}. 
Unlike the standard approach, pretrained modality-specific encoders are frozen and used exclusively for feature extraction, OneEncoder employs a single encoder to achieve greater efficiency. This differs from OneLLM, which uses a mixture of projection experts for the UP module.\newline
\indent We propose a two-step process for progressively training our framework across multiple modalities. In Step 1, we pretrain the UP using image-text data, given that this
type of data is more abundant compared to other modalities. In Step 2, which is consistent across all future modalities, we freeze the pretrained UP and train only the compact AL. This step aligns new modalities with the already aligned ones. For example, we first align audio with image and text, followed by aligning video with image, text, and audio.\newline
\indent The role of the AL is not to enhance representation but merely to project the new data into this shared space. Using OneEncoder represents a balanced compromise between alignment performance and complexity, as it minimizes the number of parameters to tune. This approach enables the integration of new modalities at a low cost and ensures good performance even in the absence of vast aligned datasets. Our contributions can be encapsulated in three key points:\newline

\begin{itemize}
    \item We introduce OneEncoder, an alignment framework designed to progressively align four modalities: image, text, audio and video.
    To achieve effective alignment without relying on vast aligned datasets, we focus on lightweight models that do not require extensive data like larger models do. We minimize the number of parameters trained in each step. In Step 1, we train only the UP. In Step 2, we train only the AL, which is even lighter than the UP because the UP is already trained and provides a well-defined shared alignment space.\newline
    
    \item We leverage lightweight framework architecture to achieve consistent alignment, even in scenarios where vast aligned datasets are unavailable.\newline
    
    \item In various downstream tasks such as classification, querying, semantic and visual question answering, our method outperforms classical methods, which trains modality-specific encoders simultaneously.
\end{itemize}

\section{Related works}
\indent The progression of large language models (LLMs) has led to their use in diverse domains beyond natural language processing, such as vision and audio. Due to their increasing power, these models are now applied to cross-modal learning tasks to align different modalities in a common semantic space. This alignment enhances representation performance by enabling the learning of one modality using information from other modalities.
\newline

\indent \textbf{Dual Modality Alignment (DMA).} Integrates representations from distinct modalities into a unified semantic space, promoting seamless interaction. Pioneering efforts, such as Flamingo~\cite{flamingo}, focus on aligning image and text modalities by incorporating visual features into LLMs with cross-attention layers, enhancing performance in vision-language tasks. Similarly, ConVIRT~\cite{zhang2022convirt} employs contrastive learning in small aligned medical datasets, while CLIP~\cite{radford2021learning} scales up contrastive learning in large aligned datasets, advancing the understanding of concepts across modalities. Additionally, ALIGN~\cite{chung2021align} addresses the challenge of noisy datasets, ensuring robust performance in real-world scenarios.\newline
\indent Building upon promising results in image-text alignment, recent works have sought to expand this approach to additional modalities such as text-audio~\cite{ristea2024cascaded,wang2023unsupervised}, image-audio~\cite{chung2022crossmodal}, text-video~\cite{jiang2022crossmodal}, and others. While these methods yield powerful representation and alignment, they suffer from two significant limitations: 1) They typically require extensive resources for training on vast aligned datasets, and 2) They are restricted to only two modalities, constraining their generalization to other modalities.
\newline

 \indent    \textbf{Multiple Modalities Alignment (MMA).} Extends the concept of DMA by synchronizing representations from more than two distinct modalities into a shared semantic space. For instance, AudioCLIP~\cite{guzhov2021audioclip} extends CLIP to handle audio alongside text and images, while ImageBInD~\cite{girdhar2023imagebind} aligns six modalities using zero-shot capabilities of vision-language models. NExT-GPT~\cite{wu2023nextgpt} enables any-to-any multimodal understanding and output but still relies on aligned datasets for training, posing challenges similar to DMA. Transitioning from heavyweight to lightweight models becomes crucial to address resource-intensive training requirements and ensure performance on diverse modalities even with limited aligned data.
\newline

\indent \textbf{Transitioning to Lightweight Models for Modalities Alignment.} Employs frozen pretrained models and modality tokens to handle multiple modalities with a single encoder, reducing the need for large aligned datasets and minimizing parameter learning~\cite{mirchandani2023large,Zhou2022,zhang2023metatransformer}. Meta-Transformer~\cite{zhang2023metatransformer} demonstrated competitive performance across 12 data modalities with a frozen visual encoder, inspiring Han et al.~\cite{han2023onellm} to propose a unified framework for aligning eight modalities using a frozen CLIP model and a single Universal Projection module (UP) with modality tokens for switching. While addressing the heavyweight nature of previous approaches, these methods struggle to integrate new modalities into existing alignments.
\newline
\indent Our work advances the lightweight framework paradigm through two key contributions: proposing an open framework that progressively aligns modalities, enabling seamless integration of new modalities with existing ones without retraining; and leveraging the lightweight nature of our approach to perform effectively even in the absence of vast aligned datasets.

\section{Method}
\label{sec:ours} 
\indent In this section, we will begin by introducing the OneEncoder architecture (Section~\ref{sec:architecture}), followed by an overview of the training phase (Section~\ref{sec:training}). 
\subsection{Model Architecture: OneEncoder}
\label{sec:architecture}
\begin{figure*}[!htbp]
    \centering
    \begin{subfigure}[b]{\textwidth}
        \centering
        \includegraphics[width=0.87\textwidth]{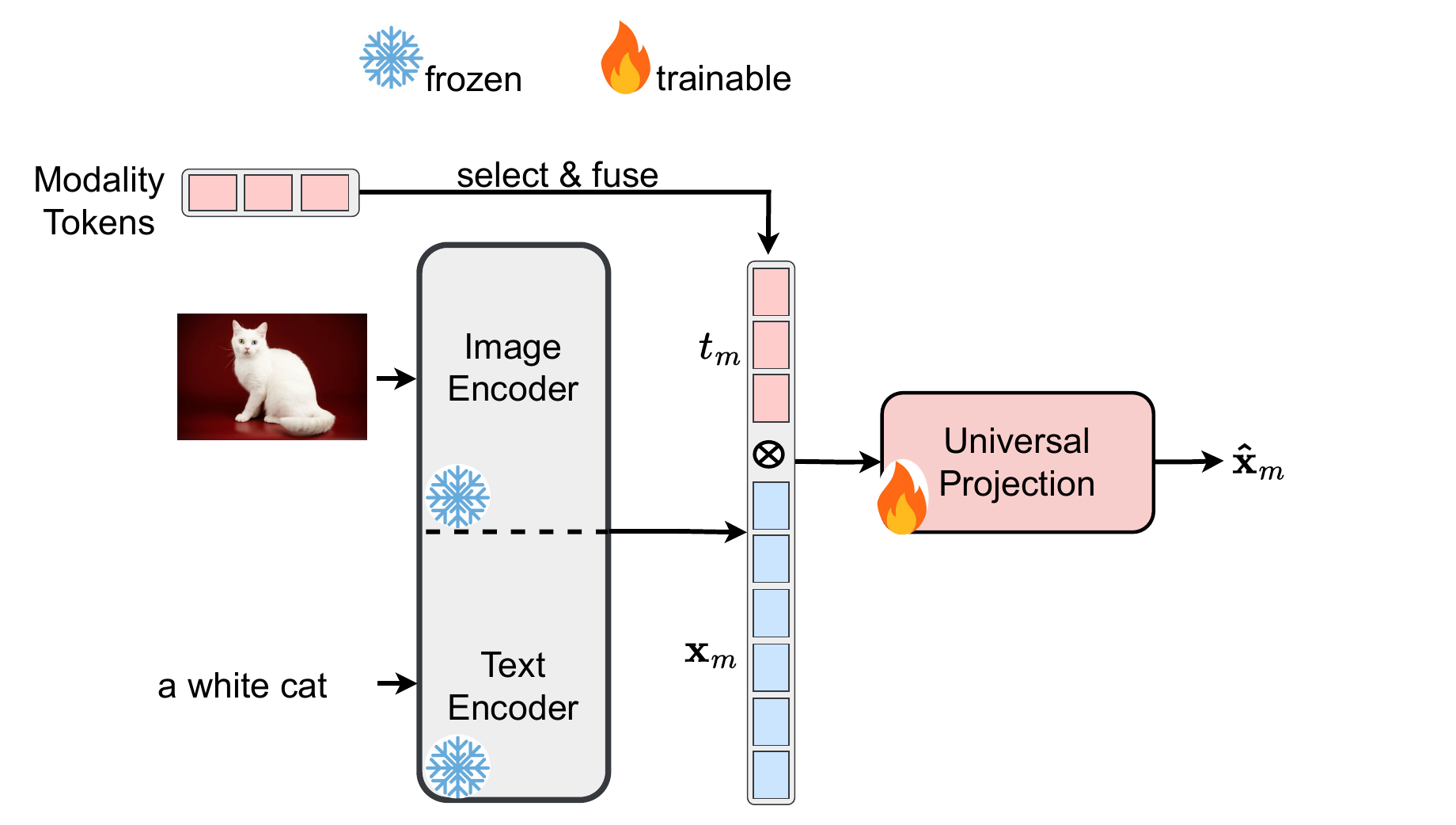} 
        \caption{\textbf{Step 1:} Training the Lightweight UP and Aligning Image-Text Modalities}
        \label{fig:step 1}
    \end{subfigure}
    \hfill
    \begin{subfigure}[b]{\textwidth}
        \centering
        \includegraphics[width=0.7\textwidth]{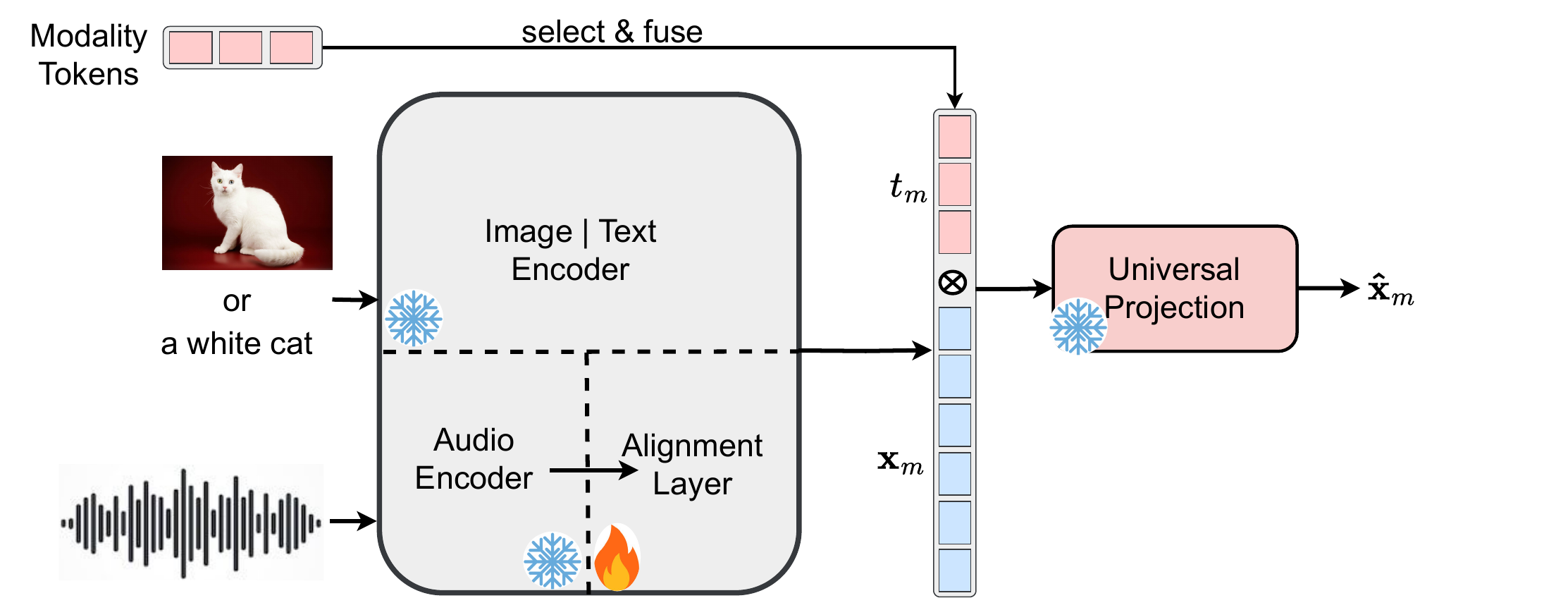} 
        \caption{\textbf{Step 2:} Freeze the Pretrained UP, Train the Compact AL, and progressively Align Audio with the Image-Text Modalities from Step 1. This process can be extended to align additional future modalities, such as video.}
        \label{fig:step 2}
    \end{subfigure}
    \caption{\textbf{OneEncoder architecture.} OneEncoder includes frozen pretrained modality-specific encoders, a Universal Projection module (UP), and an Alignment Layer (AL). In step 1, the UP, which consists of a Transformer encoder, is trained to align text and image modalities. In step 2, the pretrained UP is frozen, and the AL, composed of a multi-layer perceptron, is trained to align audio with the text and image modalities. During this step, either image or text is selected to align with audio, indirectly aligning audio with the non-selected modality. The UP fuses input ($\mathbf{x}_{\text{m}}$) and modality tokens ($\mathbf{t}_{\text{m}}$) to switch between modalities during a forward pass.}
    \label{fig:oneencoder}
\end{figure*}

Drawing from research by~\cite{mirchandani2023large,Zhou2022,zhang2023metatransformer,han2023onellm}, we capitalize on the robust modality transfer capabilities of pretrained encoders. This approach allows to leverage pretrained modality-specific models, who are trained on large modality-specific datasets, which are more readily available than large aligned datasets. Within OneEncoder, we employ ViT~\cite{Dosovitskiy2020} for image encoding, BERT~\cite{devlin2019bert} for text encoding, Wav2Vec2~\cite{baevski2020wav2vec} for audio encoding and VideoMAE~\cite{videomae} for video encoding. Each model produces an input token $\mathbf{x} \in \mathbb{R}^{L \times D}$ as its output, where $L$ represents the sequence length and $D$ denotes the token dimension. Consistent with previous  research~\cite{mirchandani2023large,Zhou2022,zhang2023metatransformer,han2023onellm}, we also maintain the parameters of these models frozen during training.
Figure~\ref{fig:oneencoder} illustrates the three primary elements comprising OneEncoder: modality-specific encoders, a Universal Projection module (UP), and an Alignment Layer (AL).\newline
\indent \textbf{Universal Projection Module (UP).} In contrast to existing methods that rely on modality-specific encoders and multiple projection experts, we introduce an encoder (UP) built using transformer layers~\cite{vaswani2017attention}. This module is designed to project any modality into a rich-common embedding space. To facilitate modality switching within the UP, we introduce learnable modality tokens $\{\mathbf{t}_{\text{m}}\}_{m \in \mathcal{M}} \in \mathbb{R}^{N \times D}$, as used in~\cite{han2023onellm}, which consists of $N$ tokens of dimension $D$, for each modality $m \in \mathcal{M}$. During the forward pass for modality $m$, we input the fusion of input tokens $\mathbf{x}_{\text{m}} \in \mathbb{R}^{L \times D}$ and modality tokens $\mathbf{t}_{\text{m}}$ into the UP module:
\begin{equation}
    \mathbf{\hat{x}}_{\text{m}} = \text{UP}(\mathbf{t}_{\text{m}} \otimes \mathbf{x}_{\text{m}}),
\end{equation}
where $\otimes$ denotes the fusion operation, which includes concatenation and addition as discussed in~\cite{wang2021you,smith2020multimodal,han2023onellm}, as well as cross-attention as detailed in~\cite{johnson2019cross}, with $\mathbf{t}_{\text{m}}$ serving as the query and $\mathbf{x}_{\text{m}}$ as the key and value.
As shown in Figure~\ref{fig:step 1}, the UP module is initially trained to align two modalities. It is then frozen on step 2 for the alignment of additional modalities, as depicted in Figure~\ref{fig:step 2}.\newline

\indent \textbf{Alignment Layer (AL).} As shown in Figure~\ref{fig:step 2}, the AL is designed to integrate new modality into the OneEncoder framework efficiently. It consists of a two-layer multi-layer perceptron (MLP). The use of a lighter model compared to the UP module allows for simpler, faster, and more cost-effective alignment of new modality. During the forward pass for a new modality $m$, the fusion of input tokens $\mathbf{x}_{\text{m}} \in \mathbb{R}^{L \times D}$, transformed by the AL component, and modality tokens $\mathbf{t}_{\text{m}}$ is fed into the frozen UP module:
\begin{equation}
    \mathbf{x}_{\text{m}} = \text{AL}(\mathbf{x}_{\text{m}})
\end{equation}
\begin{equation}
    \mathbf{\hat{x}}_{\text{m}} = \text{UP}(\mathbf{t}_{\text{m}} \otimes \mathbf{x}_{\text{m}})
\end{equation}
Step 2 can be repeated multiple times, each time a new modality needs to be aligned with the already aligned modalities.

\subsection{Training Procedure}
\label{sec:training}
\indent The OneEncoder alignment process follows a progressive two-step approach. The first step serves as the alignment initialization, while the second step can be repeated to incorporate additional modalities.
\begin{itemize}
\item \textbf{Step 1: Image-Text Alignment.} Using available aligned image-text datasets and advancements in the field~\cite{radford2021learning,chung2021align}, we train the UP module to align image and text modalities in a shared latent space. The UP's parameters are updated using the adapted InfoNCE loss~\cite{oord2018representation} for contrastive (text, image) representation learning by Zhang et al.~\cite{zhang2022convirt}.
\newline
\indent During training, we sample a minibatch of $K$ input pairs ($\hat{\mathbf{x}}_{\text{image}}^i$, $\hat{\mathbf{x}}_{\text{text}}^i$) from the dataset. The contrastive loss between image and text for each paris ($\hat{\mathbf{x}}_{\text{image}}^i$, $\hat{\mathbf{x}}_{\text{text}}^j$) in the minibatch can be formulated as follow: 
\begin{equation}
    \label{eqn:lij}
    \ell_{ij} = - \log \left(\frac{\exp(\langle \mathbf{\hat{x}}^i_{\text{image}}, \mathbf{\hat{x}}^j_{\text{text}} \rangle / \tau)}{\sum_{k=1}^K \exp(\langle \mathbf{\hat{x}}^i_{\text{image}}, \mathbf{\hat{x}}^k_{\text{text}} \rangle/\tau)}\right)
\end{equation}
The term $\langle \mathbf{\hat{x}}^i_{\text{image}}, \mathbf{\hat{x}}^j_{\text{text}} \rangle$ represents cosine similarity, with $\tau \in \mathbb{R}^+$ as a temperature parameter. This loss function preserves mutual information between true pairs through representation functions. To ensure symmetry, we introduce a similar contrastive loss from text to image:
\begin{equation}
    \label{eqn:lji}
    \ell_{ji} = - \log \left(\frac{\exp(\langle \mathbf{\hat{x}}^i_{\text{image}}, \mathbf{\hat{x}}^j_{\text{text}} \rangle / \tau)}{\sum_{k=1}^K \exp(\langle \mathbf{\hat{x}}^k_{\text{image}}, \mathbf{\hat{x}}^j_{\text{text}} \rangle/\tau)}\right)
\end{equation}
The matching pairs are situated along the diagonal of the similarity matrix ($\hat{\mathbf{x}}_{\text{image}}^i$, $\hat{\mathbf{x}}_{\text{text}}^i$), which serves as the target for the loss function:
\begin{equation}
\label{eqn:tij}
    t_{ij} =  \frac{\exp((\langle \mathbf{\hat{x}}^i_{\text{image}}, \mathbf{\hat{x}}^j_{\text{image}} \rangle + \langle \mathbf{\hat{x}}^i_{\text{text}}, \mathbf{\hat{x}}^j_{\text{text}} \rangle)  / 2\cdot \tau)}{\sum_{k=1}^K \exp((\langle \mathbf{\hat{x}}^i_{\text{image}}, \mathbf{\hat{x}}^k_{\text{image}} \rangle + \langle \mathbf{\hat{x}}^i_{\text{text}}, \mathbf{\hat{x}}^k_{\text{text}} \rangle) /2 \cdot \tau)}
\end{equation}
\begin{equation}
\label{eqn:tji}
    t_{ji} =  \frac{\exp((\langle \mathbf{\hat{x}}^i_{\text{image}}, \mathbf{\hat{x}}^j_{\text{image}} \rangle + \langle \mathbf{\hat{x}}^i_{\text{text}}, \mathbf{\hat{x}}^j_{\text{text}} \rangle)  / 2\cdot \tau)}{\sum_{k=1}^K \exp((\langle \mathbf{\hat{x}}^j_{\text{image}}, \mathbf{\hat{x}}^k_{\text{image}} \rangle + \langle \mathbf{\hat{x}}^j_{\text{text}}, \mathbf{\hat{x}}^k_{\text{text}} \rangle) /2 \cdot \tau)}
\end{equation}
The ultimate training loss $\mathcal{L}$ \eqref{eqn:cost} is computed by combining the two losses $\ell_{ij}$ and $\ell_{ji}$ and averaging them over all pairs within each minibatch.
\begin{equation}
\label{eqn:cost}
    \mathcal{L} = \frac{1}{2\cdot K}\sum_{i=1}^K \sum_{j=1}^K t_{ij}\cdot \ell_{ij} + t_{ji}\cdot \ell_{ji} 
\end{equation}

\item \textbf{Step 2: Alignment of Future Modalities.} Once the UP module is trained in Step 1, it is frozen for Step 2. In this step, a new modality $m_i$ is aligned with the already aligned image and text modalities by selecting one (either image or text) for alignment, as illustrated in Figure~\ref{fig:step 2} using the audio modality. The alignment of the selected modality ensures transitive alignment across all three modalities (image, text, and $m_i$). During this step, only the AL is trained, using the same loss function as in Step 1 (Equation~\ref{eqn:cost}) to update its parameters for consistent input to the UP module. This process is repeated whenever a new modality $m_j$ is introduced (e.g., video).
\end{itemize}

\begin{algorithm}[!ht]
\caption{Step 1: Training the Universal Projection (UP) model on the image-text modality}\label{alg:one}
\SetKwInOut{KwIn}{Input}
\SetKwInOut{KwOut}{Output}
\KwIn{image\_encoder; text\_encoder; $\mathbf{I}$: minibatch of aligned images; $\mathbf{T}$: minibatch of aligned texts; UP: transformer; $\mathcal{M} = \{\text{image, text}\}$ ;  $\{\mathbf{t}_{\text{m}}\}_{\text{m} \in \mathcal{M}} \in \mathbb{R}^{N \times D}$; $\tau$: learned temperature parameter; $\otimes$: fusion operator
}
\KwOut{Trained UP; List of aligned modalities; modality tokens
}   
    // {\small \it Freeze the pretrained encoders} \newline
    - Freeze(image\_encoder) \newline
    - Freeze(text\_encoder) \newline

    // {\small \it Extract feature representations of each modality}\newline
    - $\mathbf{X}_{\text{image}}$ = image\_encoder($\mathbf{I}$)\newline
    - $\mathbf{X}_{\text{text}}$ = text\_encoder($\mathbf{T}$)\newline

    // {\small \it Encode each modality after selection and fusion}\newline
    - $\mathbf{\hat{X}}_{\text{image}}$ = UP($\mathbf{t}_{\text{image}}$ $\otimes$ $\mathbf{X}_{\text{image}}$) \newline
    - $\mathbf{\hat{X}}_{\text{text}}$ = UP($\mathbf{t}_{\text{text}}$ $\otimes$ $\mathbf{X}_{\text{text}}$) \newline

    // {\small \it Compute Loss and Update UP Parameters, $\mathbf{t}_{\text{image}}$, and $\mathbf{t}_{\text{text}}$}\newline
    - Compute Loss using Equation~\ref{eqn:cost}: $\mathcal{L}$($\mathbf{\hat{X}}_{\text{image}}$, $\mathbf{\hat{X}}_{\text{text}}$, $\tau$)\newline
    - Update the UP parameters, $\mathbf{t}_{\text{image}}$, and $\mathbf{t}_{\text{text}}$ using an optimizer algorithm based on the computed loss.\newline

    // {\small \it Return the trained UP, list of aligned modalities $\mathcal{M}$, modality tokens}\newline
    - return UP, $\mathcal{M}$, $\{\mathbf{t}_{\text{m}}\}_{\text{m} \in \{\text{image}, \text{text}\}}$ 
\end{algorithm}
Algorithm~\ref{alg:one} provides a detailed procedure for training the UP on text-image modalities. Once trained, the UP is utilized in Algorithm~\ref{alg:two} to align a new modality, denoted as $\text{m}_2$, with the set of already aligned modalities, $\mathcal{M}$, using an intermediary modality $\text{m}_1$, where $\text{m}_1$ must be part of $\mathcal{M}$. This alignment process is achieved by training the AL to project the new modality, $\text{m}_2$, into a coherent space compatible with the UP representation. After this process, the expanded set of aligned modalities becomes $\mathcal{M} \cup \{\text{m}_2\}$. This alignment can be repeated indefinitely, allowing additional modalities to be aligned with those already in $\mathcal{M}$.\newline
\begin{algorithm}[!ht]
\caption{Step 2: Align a new modality with the previously aligned modalities}\label{alg:two}
\SetKwInOut{KwIn}{Input}
\SetKwInOut{KwOut}{Output}

\KwIn{$\text{m}_1$\_encoder; $\text{m}_2$\_encoder; $\mathbf{M}_1$: minibatch of aligned $\text{m}_1$ modality; $\mathbf{M}_2$: minibatch of aligned $\text{m}_2$ modality; UP: pretrained transformer in algorithme~\ref{alg:one}; $\mathcal{M}$: aligned modalities;  $\{\mathbf{t}_\text{m}\}_{\text{m} \in \{\text{m}_1, \text{m}_2\}} \in \mathbb{R}^{N \times D}$: modality tokens
; $\tau$: learned temperature parameter; $\otimes$: fusion operator; $\text{AL}$: Multi-layer Perceptron
}
\KwOut{Trained AL; List of aligned modalities; $\mathbf{t}_{\text{m}_2}$
}   
    // {\small \it Freeze the pretrained encoders, UP and $\text{m}_1$ modality token} \newline
    - Freeze($\text{m}_1$\_encoder) \newline
    - Freeze($\text{m}_2$\_encoder) \newline
    - Freeze(UP)\newline
    - Freeze($\mathbf{t}_{\text{m}_1}$)\newline

    // {\small \it Extract feature representations of each modality}\newline
    - $\mathbf{X}_{\text{m}_1}$ = $\text{m}_1$\_encoder($\mathbf{M}_1$)\newline
    - $\mathbf{X}_{\text{m}_2}$ = $\text{m}_2$\_encoder($\mathbf{M}_2$)\newline

    {\small \it Project feature representations with the AL}\newline
    - $\mathbf{X}_{\text{m}_1}$ = AL($\mathbf{X}_{\text{m}_1}$)\newline
    - $\mathbf{X}_{\text{m}_2}$ = AL($\mathbf{X}_{\text{m}_2}$)\newline

    // {\small \it Encode each modality after selection and fusion}\newline
    - $\mathbf{\hat{X}}_{\text{m}_1}$ = UP($\mathbf{t}_{\text{m}_1}$ $\otimes$ $\mathbf{X}_{\text{m}_1}$) \newline
    - $\mathbf{\hat{X}}_{\text{m}_2}$ = UP($\mathbf{t}_{\text{m}_2}$ $\otimes$ $\mathbf{X}_{\text{m}_2}$) \newline

    // {\small \it Compute Loss and Update AL Parameters and $\mathbf{t}_{\text{m}_2}$}\newline
    - Compute Loss using Equation~\ref{eqn:cost}: $\mathcal{L}$($\mathbf{\hat{X}}_{\text{m}_1}$, $\mathbf{\hat{X}}_{\text{m}_2}$, $\tau$)\newline
    - Update the AL parameters and $\mathbf{t}_{\text{m}_2}$ using an optimizer algorithm based on the computed loss.\newline

    // {\small \it Return the trained AL, list of aligned modalities and $\text{m}_2$} modality token\newline
    - Update list of aligned modalities: $\mathcal{M} = \mathcal{M} \cup \{\text{m}_2\}$\newline
    - return AL, $\mathcal{M}$, $\mathbf{t}_{\text{m}_2}$
\end{algorithm}

\begin{figure*}[!htbp]
    \centering
    \begin{subfigure}[b]{\textwidth}
        \centering
        \includegraphics[width=0.6\textwidth]{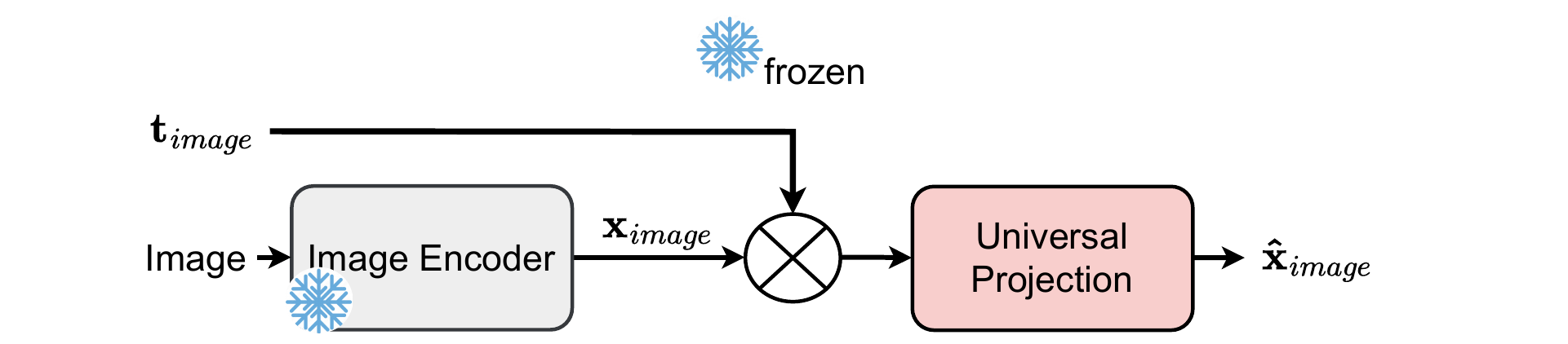} 
        \caption{Image processing}
        \label{fig:image}
    \end{subfigure}
    \hfill
    \begin{subfigure}[b]{\textwidth}
        \centering
        \includegraphics[width=0.6\textwidth]{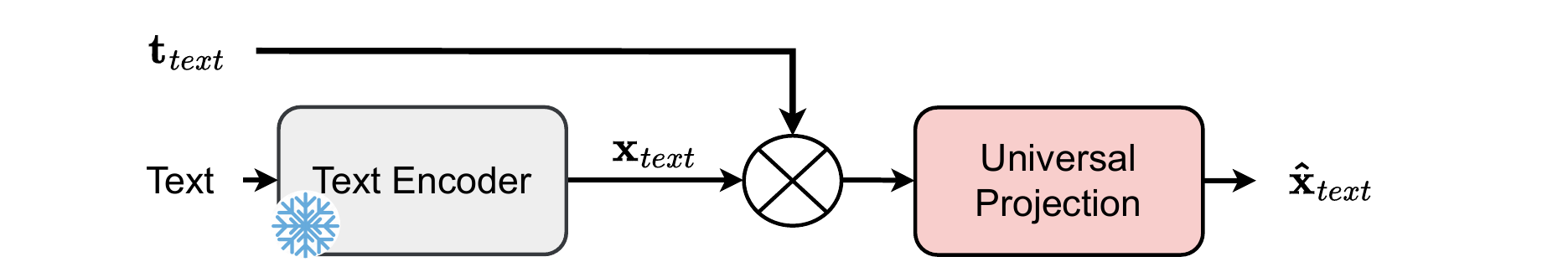} 
        \caption{Text processing}
        \label{fig:text}
    \end{subfigure}
    \hfill
    \begin{subfigure}[b]{\textwidth}
        \centering
        \includegraphics[width=0.6\textwidth]{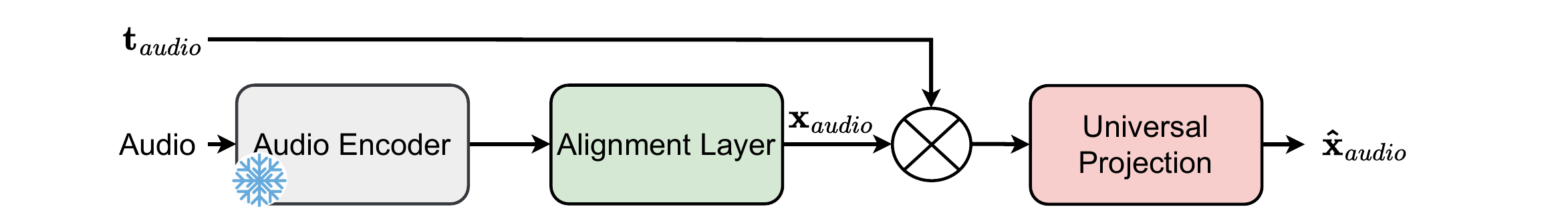} 
        \caption{Audio processing}
        \label{fig:audio}
    \end{subfigure}
    \begin{subfigure}[b]{\textwidth}
        \centering
        \includegraphics[width=0.6\textwidth]{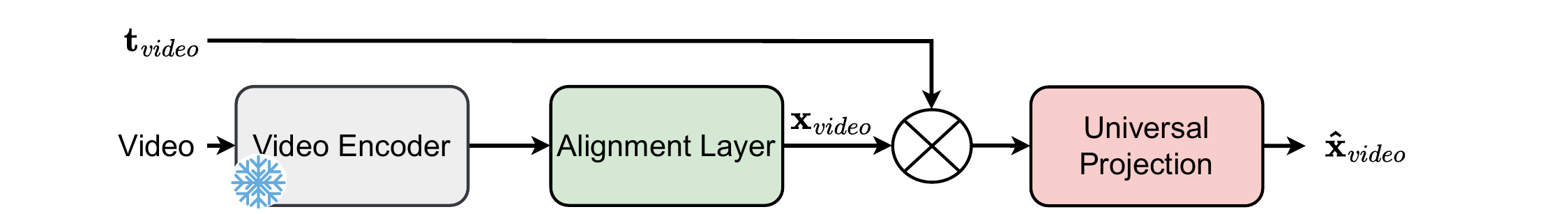} 
        \caption{Video processing}
        \label{fig:video}
    \end{subfigure}
    \caption{After training, OneEncoder can be used for various downstream tasks: in zero-shot mode by freezing the UP (Universal Projection) and AL (Alignment Layer), or fine-tuned for other tasks.}
    \label{fig:inference}
\end{figure*}

\begin{algorithm}[!ht]
\caption{Inference: Encoding a Given Modality Using Pretrained UP and AL}\label{alg:three}
\SetKwInOut{KwIn}{Input}
\SetKwInOut{KwOut}{Output}
\KwIn{m: modality of the data to be encoded; $\mathbf{M}$: minibatch of data from modality m; UP: Universal Projection model; $\text{AL}_{\text{m}}$: Alignment Layer for modality m; $\textbf{t}_{\text{m}}$: token representing modality m; m\_encoder: encoder for modality m
}
\KwOut{Encoded representation data
}   
    // {\small \it Extract feature representations}\newline
    $\mathbf{X}$ = m\_encoder($\mathbf{M}$)\newline
    
    \If{$\text{m} \notin \{\text{image}, \text{text}\}$}{
    // {\small \it Use AL for feature projection}\newline
        $\mathbf{X}$ = $\text{AL}_{\text{m}}$($\mathbf{X}$)\newline
        
    }
    
    // {\small \it Encode with the Universal Projection}\newline
    $\mathbf{\hat{X}}$ = UP($\mathbf{t}_\text{m} \otimes \mathbf{X}$)\newline

    // {\small \it Return encoded representation of input data}\newline
    return $\mathbf{\hat{X}}$
    
\end{algorithm}
In Algorithm~\ref{alg:three}, the OneEncoder framework is used to represent any modality in $\mathcal{M}$. For text and image modalities, only the UP is required, while for other modalities, both the UP and AL are necessary. Figure~\ref{fig:inference} illustrates this process, with Figures~\ref{fig:image} and~\ref{fig:text} showing how the UP handles text and image representations, and Figures~\ref{fig:audio} and~\ref{fig:video} demonstrating the use of both the UP and AL for audio and video. This approach can be applied post-training for tasks such as zero-shot inference or fine-tuning for domain-specific adaptation.

\section{EXPERIMENT}
\label{sec:experiment}
In this experiment, we aim to use OneEncoder to align four different modalities: image, text, audio, and video. Given the greater availability of datasets paired with text, we propose leveraging text as the central modality for transitive alignment. The alignment process can be summarized as follows:

\begin{enumerate}
    \item \textbf{Align Image with Text:} Train the UP using Algorithm~\ref{alg:one} on the image-text modality pair.\newline
    
    \item \textbf{Align Audio with Image and Text:} Train $\text{AL}_{\text{audio}}$ using Algorithm~\ref{alg:two} on the text-audio modality pairs.\newline
    
    \item \textbf{Align Video with Image, Text, and Audio:} Train $\text{AL}_{\text{video}}$ using Algorithm~\ref{alg:two} on the video modality pairs.\newline
    
\end{enumerate}

The order of alignment steps can be adjusted based on the availability of aligned data and the specific modalities to be aligned.

\subsection{Datasets}
\textbf{Training Datasets.} Our goal is to achieve robust performance on downstream tasks using a lightweight framework trained on a modest dataset. Following the approach of virTex~\cite{lu2020virtext} and related studies~\cite{karpathy2015deep,xu2015show}, we train the UP module on a combined dataset, which includes COCO Captions~\cite{chen2015microsoft}, Flickr30K~\cite{young2014image}, and TextCaps~\cite{sidorov2020textcaps}.\newline
To train the $\text{AL}_{\text{audio}}$, we utilize the LibriSpeech Speech Recognition Alignment (SRA)~\cite{panayotov2015librispeech} Dataset, a corpus containing approximately 1,000 hours of 16kHz recorded English speech.\newline
For the $\text{AL}_{\text{video}}$, we employ the Microsoft Research Video to Text (MSR-VTT)~\cite{xu2016msr} dataset, a large-scale resource designed for open-domain video captioning.\newline
A detailed description of all datasets used in training the OneEncoder framework is provided in Table~\ref{training_dataset}.\newline
\begin{table}[!htbp]
    \centering
    \setlength{\tabcolsep}{0.4pt} 
    \renewcommand{\arraystretch}{1.1} 
    \begin{tabular}{l|ccc|ccc|ccc|ccc}
        \hline
        \textbf{Dataset} & \textbf{Type} & \textbf{Training Size} & \textbf{Validation Size} \\
        \hline
        COCO Captions~\cite{chen2015microsoft} & text-image pairs  & 413,915  &  202,520  \\
        Flickr30K~\cite{young2014image} & text-image pairs & 158,915  & \_ \\
        TextCaps~\cite{sidorov2020textcaps} & text-image pairs & 109,765 & 15,830   \\
        SRA~\cite{panayotov2015librispeech} & text-audio pairs & 281,241 & 5,559 \\
        MSR-VTT~\cite{xu2016msr} & text-video pairs & 6,513  & 497 \\
        \hline
    \end{tabular}
    
    \caption{
        Training datasets
    }
   \label{training_dataset}
\end{table}
\textbf{Validation Datasets.} For validating OneEncoder, we use various datasets, tailored either for specific modality-based validation (e.g., classification tasks) or cross-modal validation (e.g., zero-shot tasks). A comprehensive description of the datasets used for validation is provided in Table~\ref{validation_dataset}.
\begin{table}[!htbp]
    \centering
        \setlength{\tabcolsep}{0.4pt} 
    \renewcommand{\arraystretch}{1.1} 
    \begin{tabular}{l|ccc|ccc|ccc|ccc}
        \hline
        \textbf{Dataset} & \textbf{Dataset Type} & \textbf{Training Size} & \textbf{Validation Size} \\
        \hline
        CIFAR-10~\cite{cifar10} & image  & 50,000 & 10,000 \\
        Oxford-IIIT Pets~\cite{parkhi2012cats} & image & 3,680 & 3,669 \\
        CIFAR-100~\cite{cifar100} & image & 50,000 & 10,000 \\
        Caltech 101~\cite{fei2004learning} & image & 7,659 & 3,060 \\
        Tiny ImageNet~\cite{le2015tiny} & image & 100,000 & 10,000 \\
        SST-2~\cite{socher2013recursive} & text & 67,349 & 872 \\
        TREC~\cite{voorhees1999trec} & text & 5,452 & 500 \\
        Emotion~\cite{saravia2018carer} & text & 16,000 & 2,000 \\
        GTZAN~\cite{tzanetakis2002musical}  & audio & 1,000 & \_ \\
        UrbanSound8K~\cite{salamon2014dataset}  & audio & 7,980 & 1,022 \\
        ESC-50~\cite{piczak2015esc} & audio  & 1,600 & 400  \\
        MSVD~\cite{chen2011collecting} & video-text & 48,779 & 4,291 \\
        LSMDC~\cite{liu2020usehavevideoretrieval} & video-text & 118,081 & \_ \\
        DAQUAR~\cite{malinowski2014multi} & image-text & 6794 & 5,673 \\
        \hline
    \end{tabular}
    
    \caption{
        Validation datasets
    }
   \label{validation_dataset}
\end{table}

\subsection{Implementation Details}
\textbf{Architecture.} The pretrained encoders for each modality are as follows: ViT-base~\cite{orton2020vision} with 86M parameters for images, BERT-base~\cite{devlin2019bert} with 110M parameters for text, Wav2Vec~\cite{baevski2020wav2vec} with 317M parameters for audio and VideoMAE-base~\cite{videomae} with 94.2M for video. Additionally, the UP encoder consists of four Transformer block with 4M parameters, while the AL comprises a multi-layer perceptron with 65,792 parameters. The size
of modality tokens for each modality is $\mathbb R^{1 \times 768}$.\newline

\indent \textbf{Training Details.} We use the AdamW optimizer~\cite{loshchilov2019decoupled} with a learning rate of $0.001$, $\beta_1=0.9$, $\beta_2=0.95$, and a weight decay of $0.001$. For step 1, we train to align image-text pairs, updating only the UP parameters, on a single A100 GPU for 500 epochs with a batch size of 512. For step 2, to align other modalities (audio and video) , we freeze the pretrained UP from step 1, and train only the $\text{AL}_{\text{m}}, \text{m} \in \{\text{audio, video}\}$  for 100 epochs, using the same parameters as in step 1 with a batch size of 64.\newline
\indent We trained two OneEncoder models, each utilizing a different fusion operation: addition and scaled dot product attention~\cite{vaswani2017attention}. For simplicity, we refer to the model using addition as~\textbf{OneEncoder-1}, and the model using scaled dot product attention as~\textbf{OneEncoder-2}.\newline
\indent Our objective is not to achieve state-of-the-art results, which typically demand resource-intensive architectures and extensive hyperparameter tuning. Instead, we aim to explore the behavior of frozen versus non-frozen modality-specific encoders. Specifically, we seek to demonstrate that using frozen encoders within our OneEncoder framework can notably enhance performance and, in many cases, yield better representations for downstream tasks. For a fair comparison, we refer to the baseline approach, which involves training modality-specific encoders, as the {\textbf{Base}} framework.

\subsection{Quantitative Evaluation}
\subsubsection{UP Validation Following Image-Text Modalities Training}
\label{train_up}
After training the UP module on a combined dataset of COCO Captions, Flickr30K, and TextCaps, we validate the OneEncoder framework by benchmarking it against the baseline CLIP model~\cite{radford2021learning}. In our method, the pretrained ViT and BERT models remain frozen during training, with only the UP module's 4M parameters being updated. In contrast, the baseline requires training all 196M parameters of the ViT and BERT models. For specific tasks, we employ pretrained models: ResNet-18~\cite{he2015deepresiduallearningimage}, EfficientNet-B0~\cite{tan2020efficientnetrethinkingmodelscaling}, and Swin Transformer~\cite{liu2021swintransformerhierarchicalvision} for image processing, and RoBERTa~\cite{roberta}, DistilBERT~\cite{sanh2020distilbertdistilledversionbert}, and XLNet~\cite{yang2020xlnetgeneralizedautoregressivepretraining} for text processing. 

We encode each modality using Algorithm~\ref{alg:three} within the OneEncoder framework and evaluate the performance on various classification tasks.\newline

\paragraph{Zero-shot Classification} is a task where a model, trained on labeled images, can classify new images from previously unseen classes. It validates the model's generalization capability and assesses semantic understanding and transfer learning. Using the CLIP approach, we transform labels into text descriptions (\textit{"A photo of a \{label}\}."), encode them with a pretrained model, compute cosine similarity with image embeddings, and use softmax to determine class probabilities.\newline
\begin{table}[!ht]
    \centering
    \resizebox{0.47\textwidth}{!}{
    \setlength{\tabcolsep}{2pt} 
    \renewcommand{\arraystretch}{1.5} 
    \begin{tabular}{l|ll|lll|ll|lll|lll}
        \hline
        \textbf{model}  &  \textbf{CIFAR-10} & \textbf{Oxford-IIIT Pets} & \textbf{CIFAR-100} & \textbf{Caltech-101} & \textbf{Tiny ImageNet} \\
        \hline
        CLIP & 62.12 &  58.27 & 53.06 & 52.17 & 47.15 \\
        \hline
        OneEncoder-1 & \textbf{78.15} & \textbf{69.23} & \textbf{58.18} & \textbf{56.20} & \textbf{52.27} \\
        OneEncoder-2 & 74.70 & 68.98 & 57.15 & 54.12  & 51.12 \\
        \hline
    \end{tabular}
    }
    \caption{
    Image-Text Alignment Validation: Zero-shot image classification is used to assess the alignment accuracy (\%) across five benchmark datasets with varying class counts, providing a measure of the relevance and effectiveness of the image-text alignment.
    }
   \label{table:zero_shot}
\end{table}
Zero-shot image classification obviates the need for retraining pretrained models on target datasets, evaluating their ability to generalize to unseen classes. It underscores the importance of the aligned latent space. Results in Table~\ref{table:zero_shot} highlight superior performance of OneEncoder (OneEncoder-1, OneEncoder-2) over the baseline (CLIP) across all datasets, suggesting that training large modality-specific encoders may not always be optimal, as demonstrated by the effectiveness of the lightweight OneEncoder framework.
\newline

\paragraph{{Linear Classification and Fine-Tuning}} involve adding a linear classifier to a pretrained model, freezing the pretrained weights and training only the linear classifier for linear classification, while training both the pretrained model and the linear classifier for fine-tuning. Linear classification allows for the assessment of the quality of the extracted features from the pretrained model, while fine-tuning simulates the practical use of pretrained weights. In OneEncoder, we always freeze the modality-specific encoders; in the fine-tuning task, we train only the UP for image and text datasets. In each case (Linear Classification and Fine-Tuning), we train models for 100 epochs without using any data augmentation strategy.\newline
\begin{table*}[!ht]
    \centering
    \resizebox{\textwidth}{!}{%
    \begin{tabular}{l|lllll|lll}
        \hline
        \multicolumn{9}{c}{\textbf{Linear Classification}} \\
        \hline
        \textbf{Model} & \multicolumn{5}{c|}{\textbf{Image Classification}} & \multicolumn{3}{c}{\textbf{Text Classification}}  \\
        & CIFAR-10 & Oxford-IIIT Pets & CIFAR-100 & Caltech-101 & Tiny ImageNet & SST-2 & TREC & Emotion  \\
        \hline
       ResNet-18 & 89.15 & 84.98 &  68.10 & 63.45 & 59.11 & \_ & \_ & \_ \\
       EfficientNet-B0 & 89.87 & 85.12 &  70.15 & 64.87 & 60.27 & \_ & \_ & \_\\
       Swin Transformer & 90.17 & 86.05 &  \textbf{71.12} & 65.10 & \textbf{62.30} & \_ & \_ & \_\\
       RoberTa & \_ & \_ &  \_ & \_ & \_ & 76.04 & 77.34 & 59.06  \\
       DistilBERT& \_ & \_ &  \_ & \_ & \_ & 77.15 & 76.14 & 68.11  \\
       XLNet & \_ & \_ &  \_ & \_ & \_ & 79.27 & 78.11 & 60.10  \\
       CLIP & 81.21 & 78.16 &  60.12 & 60.14 & 58.14 & 80.15 & 78.24 & 60.23  \\
       OneEncoder-1 & \textbf{90.16} & 86.23 &  70.10 & \textbf{68.23} & 62.12 & \textbf{82.12} & \textbf{79.10} & \textbf{63.09} \\
       OneEncoder-2 & 89.18 & \textbf{86.78} &  68.27 & 65.05 & 60.10 & 80.87 & 78.06 & 61.89  \\
        \hline
    \end{tabular}
    }
    \resizebox{\textwidth}{!}{
    \begin{tabular}{l|lllll|llll}
        \hline
        \multicolumn{9}{c}{\textbf{Fine-Tuning}}\\
        \hline
        \textbf{Model} & \multicolumn{5}{c|}{\textbf{Image Classification}} & \multicolumn{3}{c}{\textbf{Text Classification}}  \\
        & CIFAR-10 & Oxford-IIIT Pets & CIFAR-100 & Caltech-101 & Tiny ImageNet & SST-2 & TREC & Emotion  \\
        \hline
        ResNet-18 & 93.23 & 90.19 &  82.37 & 78.12 & 67.89 & \_ & \_ & \_ \\
        EfficientNet-B0 & 94.56 & 92.23 &  80.11 & 79.98 & 68.10 & \_ & \_ & \_\\
        Swin Transformer & 95.27 & 92.11 &  \textbf{82.02} & 79.15 & 69.09 & \_ & \_ & \_\\
       RoberTa & \_ & \_ &  \_ & \_ & \_ & 83.24 & 85.45 & 66.13  \\
       DistilBERT & \_ & \_ & \_ & \_ & \_ & 82.56 & 83.27 & 63.15  \\
       XLNet & \_ & \_ & \_ & \_ & \_ & 84.72 & 85.67 & 64.11  \\
        CLIP & 86.76 & 81.90 &  70.87 & 69.67 & 60.15 & 85.15 & 84.24 & 64.56  \\
        OneEncoder-1 & \textbf{96.01} & 92.32 & 81.10 & \textbf{80.11} & 69.12 & \textbf{86.11} & \textbf{86.12} & \textbf{67.12}  \\
        OneEncoder-2 & 95.98 & \textbf{93.12} &  80.21 & 78.23 & \textbf{69.15} & 85.12 & 86.00 & 66.78  \\
        \hline
    \end{tabular}
    }
    \caption{
            Linear classification and fine-tuning accuracy (\%) across various benchmark datasets in image and text modalities. Linear classification involves training only a linear classifier while keeping all pretrained models frozen. Fine-tuning entails training both the pretrained models and the linear classifier. For OneEncoder models (OneEncoder-1 and OneEncoder-2), only the 
            UP (Universal Projection) component is trained during fine-tuning, with modality-specific encoders remaining frozen. In contrast, CLIP and other baseline models (ResNet-18, EfficientNet-B0, SWin Transformer, RoberTa, DistilBERT and XLNet) are retrained during fine-tuning.
            }
   \label{table:classification}
\end{table*}
The results presented in Table~\ref{table:classification} demonstrate the performance of various models on image and text classification tasks using two training strategies: linear classification and fine-tuning. These approaches allow us to evaluate the models' ability to generalize to new data, providing a comprehensive comparison between OneEncoder, CLIP, and other baselines.

In image classification, OneEncoder consistently outperforms the CLIP model, which uses CLIP-ViT on image datasets. For linear classification, OneEncoder-1 achieves the highest accuracy on CIFAR-10 (90.16\%), Oxford-IIIT Pets (86.23\%), and Caltech-101 (68.23\%), closely rivaling Swin Transformer, which leads in CIFAR-100 (71.12\%) and Tiny ImageNet (62.30\%). This highlights the efficiency of OneEncoder, especially considering that it only updates the 4M parameters of the UP module, unlike CLIP, which retrains its larger 196M parameters.

In text classification tasks, where CLIP-BERT is used as the baseline for CLIP, OneEncoder again demonstrates superior performance. OneEncoder-1 achieves the best results across all datasets: SST-2 (82.12\%), TREC (79.10\%), and Emotion (63.09\%) in the linear classification setup. This shows its robust ability to handle diverse text modalities, outperforming specialized models like RoBERTa, DistilBERT, and XLNet.

The fine-tuning results further emphasize the effectiveness of OneEncoder. For image classification, OneEncoder-1 delivers the highest accuracy on CIFAR-10 (96.01\%), Oxford-IIIT Pets (92.32\%), and Caltech-101 (80.11\%), while also performing competitively on Tiny ImageNet (69.12\%), narrowly surpassed by Swin Transformer. In text classification, OneEncoder-1 achieves the best performance on SST-2 (86.11\%), TREC (86.12\%), and Emotion (67.12\%), surpassing the fine-tuned CLIP-BERT and other text-specific models.

Overall, the results illustrate that OneEncoder, with its efficient training approach and minimal parameter updates, outperforms CLIP and other models in both image and text tasks, demonstrating its superior generalization and adaptability across multiple modalities.\newline

\subsubsection{$\text{AL}_{\text{audio}}$ Validation Following Text-Audio Modalities Training}
\label{sec:audio}
After training the UP on image-text modalities, it is frozen and then used for aligning other modalities. Specifically, for audio alignment, only the $\text{AL}_{\text{audio}}$) with 65,792 parameters is trained within the OneEncoder framework. This process uses a text-audio modality dataset and follows Algorithm~\ref{alg:two} on the ROCO dataset. For comparison, we also train AudioCLIP, an extended version of CLIP that aligns image, text, and audio using ViT for images, BERT for text, and Wav2Vec for audio, with a total of 513M parameters to tune.\newline
\begin{table}[!ht]
    \centering
    \resizebox{0.47\textwidth}{!}{
    \setlength{\tabcolsep}{2pt} 
    \renewcommand{\arraystretch}{1.5} 
    \begin{tabular}{l|lll|lll|lll}
        \hline
        \textbf{Model} & \multicolumn{3}{c|}{\textbf{AudioSet}} & \multicolumn{3}{c|}{\textbf{UrbanSound8K}} & \multicolumn{3}{c}{\textbf{ESC-50}} \\
        & P@1 & R@1 & mAP & P@1 & R@1 & mAP & P@1 & R@1 & mAP \\
        \hline
        AudioCLIP & 4.27 & 75.37 &  27.12 & 40.10 & 45.11 & 78.27 & \textbf{48.90} & 78.21 & 75.12 \\
        \hline
        OneEncoder-1 & \textbf{5.37} & \textbf{76.10} &  \textbf{28.37} & \textbf{41.11} & \textbf{46.12} & \textbf{79.65} & 47.98 & \textbf{80.12} & \textbf{75.57} \\
        OneEncoder-2 & 5.10 & 76.06 &  28.10 & 40.89 & 45.78 & 79.23 & 47.87 & 78.12 & 74.98 \\
        \hline
    \end{tabular}
    }

    \caption{
            Performance metrics for text-audio retrieval tasks on the AudioSet, UrbanSound8K, and ESC-50 datasets. The evaluation includes Top-1 Precision (P@1), Top-1 Recall (R@1), and mean Average Precision (mAP) for the models: AudioCLIP, OneEncoder-1, and OneEncoder-2.
            }
   \label{table:audio_retrieval}
\end{table}
Table~\ref{table:audio_retrieval} compares the performance of AudioCLIP and OneEncoder models (OneEncoder-1 and OneEncoder-2) in text-audio retrieval. This task validates the alignment between text and audio. Evaluated using Top-1 Precision/Recall (P@1, R@1) and mean Average Precision (mAP), OneEncoder consistently outperforms AudioCLIP across all datasets. This highlights OneEncoder's efficient latent space and its ability to handle cross-modal retrieval effectively. Unlike AudioCLIP, which requires extensive encoder training, OneEncoder achieves superior results with a lightweight framework, demonstrating its robustness with minimal dataset-specific training.\newline
\begin{table}[!ht]
    \centering
    \resizebox{0.47\textwidth}{!}{
    \setlength{\tabcolsep}{2pt} 
    \renewcommand{\arraystretch}{1.5} 
    \begin{tabular}{l|ll|lll|ll|lll|lll}
        \hline 
        model  &  CIFAR-10 & Oxford-IIIT Pets & CIFAR-100 & Caltech-101 & Tiny ImageNet \\
        \hline
        AudioCLIP & 61.28 &  58.15 & 52.27 & 51.10 & 46.04 \\
        \hline
        OneEncoder-1 & \textbf{77.01} & \textbf{69.02} & \textbf{56.07} & \textbf{55.37} & \textbf{50.18} \\
        OneEncoder-2 & 74.07 & 66.56 & 55.18 & 53.11  & 50.06 \\
        \hline

    \end{tabular}
    }
    \caption{
    Image-Audio Alignment Validation: Zero-shot image classification is used to assess the alignment accuracy (\%) across five benchmark datasets with varying class counts, providing a measure of the relevance and effectiveness of the image-audio alignment.
    }
   \label{table:zero_shot_audio}
\end{table}
To validate transitive alignment between audio and image, we apply the zero-shot classification method as described in Section~\ref{train_up}, replacing text descriptions ("A photo of a \{label\}.") with corresponding audio. Comparing Table~\ref{table:zero_shot_audio} with Table~\ref{table:zero_shot}, which uses text descriptions, demonstrates that the OneEncoder framework maintains strong alignment between image and audio, even without direct image-audio alignment. This approach is more efficient and powerful than the resource-intensive AudioCLIP, offering a cost-effective solution with superior performance.
\begin{table}[!ht]
    \centering
    \begin{tabular}{l|lll|lll}
        \hline 
        model  &  UrbanSound8K & ESC-50  \\
        \hline
        Piczak-CNN~\cite{piczak2015environmental} & 73.70 & 64.50 \\
        SB-CNN~\cite{salamon2017deep} & 79.00 & \_ \\
        ESResNet~\cite{guzhov2021esresnet} & 85.42  & 91.50 \\
        AST~\cite{gong2021ast} & \_ & 95.60 \\
        ERANN~\cite{verbitskiy2022eranns} & \_ & 96.10\\
        AudioCLIP & 88.32 &  96.12  \\
        \hline
        OneEncoder-1 & \text{89.23} & 96.87  \\
        OneEncoder-2 & 88.86 & \textbf{97.02}  \\
        \hline

    \end{tabular}
    \caption{
       Representation learning evaluation using fine-tuning on the UrbanSound8K and ESC-50 datasets.
    }
   \label{table:fine-tuning}
\end{table}
For representation learning model validation, we fine-tune the models on the UrbanSound8K and ESC-50 datasets. Unlike AudioCLIP, which requires retraining all Wav2Vec parameters, OneEncoder only fine-tunes the UP and the ($\text{AL}_{\text{audio}}$) for 100 epochs. Table~\ref{table:fine-tuning} shows that OneEncoder-1 and OneEncoder-2 outperform AudioCLIP on both datasets, with OneEncoder-2 achieving the highest accuracy on ESC-50 (97.02\%) and OneEncoder-1 leading on UrbanSound8K (89.23\%). This demonstrates the efficiency of the OneEncoder framework, achieving superior performance with fewer retrained parameters compared to the more resource-intensive AudioCLIP. These results underscore the robustness of OneEncoder for fine-tuned representation learning across diverse audio classification tasks.

\subsubsection{$\text{AL}_{\text{video}}$ Validation Following Text-Video Modalities Training}
After aligning the audio with both image and text modalities (see Section~\ref{sec:audio}), we further integrate the video modality and align it with image, text, and audio. This alignment is performed using Algorithm~\ref{alg:two}, following a similar approach as in audio alignment, where only the $\text{AL}_{video}$ is trained while keeping the UP frozen. The OneEncoder model is trained for 100 epochs on the MSR-VTT dataset, utilizing the text modality to align with the video modality, which consequently aligns the audio and image modalities with the video through transitive alignment.\newline
For evaluating OneEncoder in the context of text-video alignment, we benchmark its performance against X-CLIP~\cite{ma2022xclipendtoendmultigrainedcontrastive}, an extended version of CLIP designed for text-video alignment. 

\begin{table}[!ht]
    \centering
    \resizebox{0.5\textwidth}{!}{
    \setlength{\tabcolsep}{2pt} 
    \renewcommand{\arraystretch}{1.5} 
    \begin{tabular}{l|lll|lll}
        \hline
        \multicolumn{7}{c}{\textbf{Retrieval performance comparison on MSVD}} \\
        \hline
        \textbf{Model} & \multicolumn{3}{c|}{\textbf{Text-to-Video}} & \multicolumn{3}{c}{\textbf{Video-to-Text}}  \\
        & R@1$\uparrow$ & R@5$\uparrow$ & MnR$\downarrow$ & R@1$\uparrow$ & R@5$\uparrow$ & MnR$\downarrow$  \\
        \hline
        Multi Cues \cite{hendricks2017localizing} & 20.3 & 47.8 & - & - & - & - \\
        CE \cite{chen2020simple} & 19.8 & 49.0 & - & - & - & - \\
        SSB \cite{chen2020improved} & 28.4 & 60.0 & - & - & - & - \\
        NoiseE \cite{amrani2020noise} & 20.3 & 49.0 & - & - & - & - \\
        CLIP-straight \cite{radford2021learning} & 37.0 & 64.1 & - & 59.9 & 85.2 & - \\
        Frozen \cite{bain2021frozen} & 33.7 & 64.7 & - & - & - & - \\
        TT-CE+ \cite{gabeur2020multimodal} & 25.4 & 56.9 & - & 27.1 & 55.3 & - \\
        CLIP4Clip-MeanP (ViT-B/32) \cite{luo2021clip4clip} & 46.2 & 76.1 & 10.0 & 56.6 & 79.7 & 7.6 \\
        CLIP4Clip-seqTransf (ViT-B/32) \cite{luo2021clip4clip} & 45.2 & 75.5 & 10.3 & 62.0 & 87.3 & 4.3 \\
        CLIP4Clip-MeanP (ViT-B/16) \cite{luo2021clip4clip} & 47.3 & 77.7 & 9.1 & 62.9 & 87.2 & 4.2 \\
        CLIP4Clip-seqTransf (ViT-B/16) \cite{luo2021clip4clip} & 47.2 & 77.7 & 9.1 & 63.2 & 87.2 & 4.2 \\
        X-CLIP (ViT-B/32) & 47.1 & 77.8 & 9.5 & 60.9 & 87.8 & 4.7 \\
        X-CLIP (ViT-B/16) & \textbf{50.4} & 80.6 & 8.4 & \textbf{66.8} & 90.4 & 4.2 \\
        \hline
        OneEncoder-1 & 49.21 & \textbf{80.76} & \textbf{7.98} & 65.89 & \textbf{91.62} & \textbf{3.98} \\
        OneEncoder-2 & 47.02 &79.27 & 8.88 & 65.23 & 89.78 & 4.65 \\
        \hline
    \end{tabular}
    }
    \vspace{0.5cm} 
    \resizebox{0.5\textwidth}{!}{
    \setlength{\tabcolsep}{2pt} 
    \renewcommand{\arraystretch}{1.5} 
    \begin{tabular}{l|lll|lll}
        \hline
        \multicolumn{7}{c}{\textbf{Retrieval performance comparison on LSMDC}} \\
        \hline
        \textbf{Model} & \multicolumn{3}{c|}{\textbf{Text-to-Video}} & \multicolumn{3}{c}{\textbf{Video-to-Text}}  \\
        & R@1$\uparrow$ & R@5$\uparrow$ & MnR$\downarrow$ & R@1$\uparrow$ & R@5$\uparrow$ & MnR$\downarrow$  \\
        \hline
        CT-SAN \cite{dong2019dual} & 5.1 & 16.3 & - & - & - & - \\
        JSFusion \cite{dong2019dual} & 9.1 & 21.2 & - & 12.3 & 28.6 & - \\
        CE \cite{chen2020simple} & 11.2 & 26.9 & 96.8 & - & - & - \\
        MMT \cite{gabeur2020multimodal} & 12.9 & 29.9 & 75.0 & - & - & - \\
        NoiseE \cite{amrani2020noise} & 6.4 & 19.8 & - & - & - & - \\
        CLIP-straight \cite{radford2021learning} & 11.3 & 22.7 & - & 6.8 & 16.4 & - \\
        MDMMT \cite{dzabraev2021mdmmt} & 18.8 & 38.5 & 58.0 & - & - & - \\
        Frozen \cite{bain2021frozen} & 15.0 & 30.8 & - & - & - & - \\
        HiT \cite{bertasius2021space} & 14.0 & 31.2 & - & - & - & - \\
        TT-CE+ \cite{gabeur2020multimodal} & 17.2 & 36.5 & - & 17.5 & 36.0 & - \\
        CLIP4Clip-MeanP (ViT-B/32) \cite{luo2021clip4clip} & 20.7 & 38.9 & 65.3 & 20.6 & 39.4 & 56.7 \\
        CLIP4Clip-seqTransf (ViT-B/32) \cite{luo2021clip4clip} & 22.6 & 41.0 & 61.0 & 20.8 & 39.0 & 54.2 \\
        CLIP4Clip-MeanP (ViT-B/16) \cite{luo2021clip4clip} & 23.5 & 43.2 & 54.8 & 22.6 & 50.5 & 50.3 \\
        CLIP4Clip-seqTransf (ViT-B/16) \cite{luo2021clip4clip} & 23.5 & 45.2 & 51.6 & 23.2 & 42.4 & 47.4 \\
        X-CLIP (ViT-B/32) & 23.3 & 43.0 & 56.0 & 22.5 & 42.2 & 50.7 \\
        X-CLIP (ViT-B/16) & 26.1 & \textbf{48.4} & 46.7 & 26.9 & 46.2 & \textbf{41.9} \\
        \hline
        OneEncoder-1 & \textbf{26.12} & 48.23 & \textbf{46.11} & \textbf{27.01} & \textbf{46.67} & 42.3 \\
        OneEncoder-2 & 25.32 & 46.76 & 50.19 & 25.67 & 44.15 & 42.10 \\
        \hline
    \end{tabular}
    }
    \caption{Comparison of text and video retrieval performance on the MSVD and LSMDC datasets. The reported metrics are Recall at Rank 1 (R@1) and Rank 5 (R@5), where higher values indicate better performance, and Mean Rank (MnR), where lower values are better.}
   \label{table:retrieval_comparison}
\end{table}
Results on Table~\ref{table:retrieval_comparison} demonstrate the superior performance of OneEncoder in aligning text and video across both MSVD and LSMDC datasets. On MSVD, OneEncoder-1 outperforms all models with a notable Recall at rank 5 (R@5) of 80.76 and Mean Rank (MnR) of 7.98 in text-to-video retrieval. Similarly, in video-to-text retrieval, it achieves the best R@5 score (91.62) and the lowest MnR (3.98), surpassing strong baselines like CLIP4Clip and X-CLIP.

\begin{table}[!ht]
    \centering
    \resizebox{0.5\textwidth}{!}{
  
    \begin{tabular}{l|lll|lll}
        \hline
        \multicolumn{7}{c}{\textbf{Retrieval performance comparison on MSVD}} \\
        \hline
        \textbf{Model} & \multicolumn{3}{c|}{\textbf{Audio-to-Video}} & \multicolumn{3}{c}{\textbf{Video-to-Audio}}  \\
        & R@1$\uparrow$ & R@5$\uparrow$ & MnR$\downarrow$ & R@1$\uparrow$ & R@5$\uparrow$ & MnR$\downarrow$  \\
        \hline
        OneEncoder-1 & 46.34 & \textbf{78.45} & \textbf{8.13} & 63.43 & \textbf{89.12} & \textbf{4.15} \\
        OneEncoder-2 & \textbf{46.78} & 77.32 & 8.97 & \textbf{63.78} & 87.78 & 4.98 \\
        \hline
    \end{tabular}
    }
    \vspace{0.5cm} 
    \resizebox{0.5\textwidth}{!}{

    \begin{tabular}{l|lll|lll}
        \hline
        \multicolumn{7}{c}{\textbf{Retrieval performance comparison on LSMDC}} \\
        \hline
        \textbf{Model} & \multicolumn{3}{c|}{\textbf{Audio-to-Video}} & \multicolumn{3}{c}{\textbf{Video-to-Audio}}  \\
        & R@1$\uparrow$ & R@5$\uparrow$ & MnR$\downarrow$ & R@1$\uparrow$ & R@5$\uparrow$ & MnR$\downarrow$  \\
        \hline
        OneEncoder-1 & \textbf{24.37} & \textbf{46.32} & \textbf{47.32} & \textbf{25.23} & \textbf{44.56} & \textbf{44.20} \\
        OneEncoder-2 & 23.32 & 43.13 & 49.32 & 23.21 & 42.12 & 46.57 \\
        \hline
    \end{tabular}
    }
    \caption{Audio and video retrieval performance on the MSVD and LSMDC datasets. The reported metrics are Recall at Rank 1 (R@1) and Rank 5 (R@5), where higher values indicate better performance, and Mean Rank (MnR), where lower values are better.}
   \label{table:retrieval_comparison_audio}
\end{table}
To validate the transitive alignment between audio and video, we convert each text description into audio and perform audio-video retrieval to assess alignment. Table~\ref{table:retrieval_comparison_audio} compares these results with those in Table~\ref{table:retrieval_comparison}, demonstrating successful audio-video alignment. This confirms the effectiveness of the progressive alignment process, which requires minimal computational resources while maintaining the strong performance of the OneEncoder framework.\newline

\indent These quantitative results highlight the efficiency of OneEncoder's lightweight architecture, which achieves high-performance alignment across multiple modalities at low computational cost. This framework can be effectively applied post-training in various tasks such as classification and retrieval, or fine-tuned for specific domains to enhance representation quality.

\subsection{Qualitative Analysis}
\begin{figure*}[!htbp]
      \centering
      \includegraphics[width=0.87\textwidth]{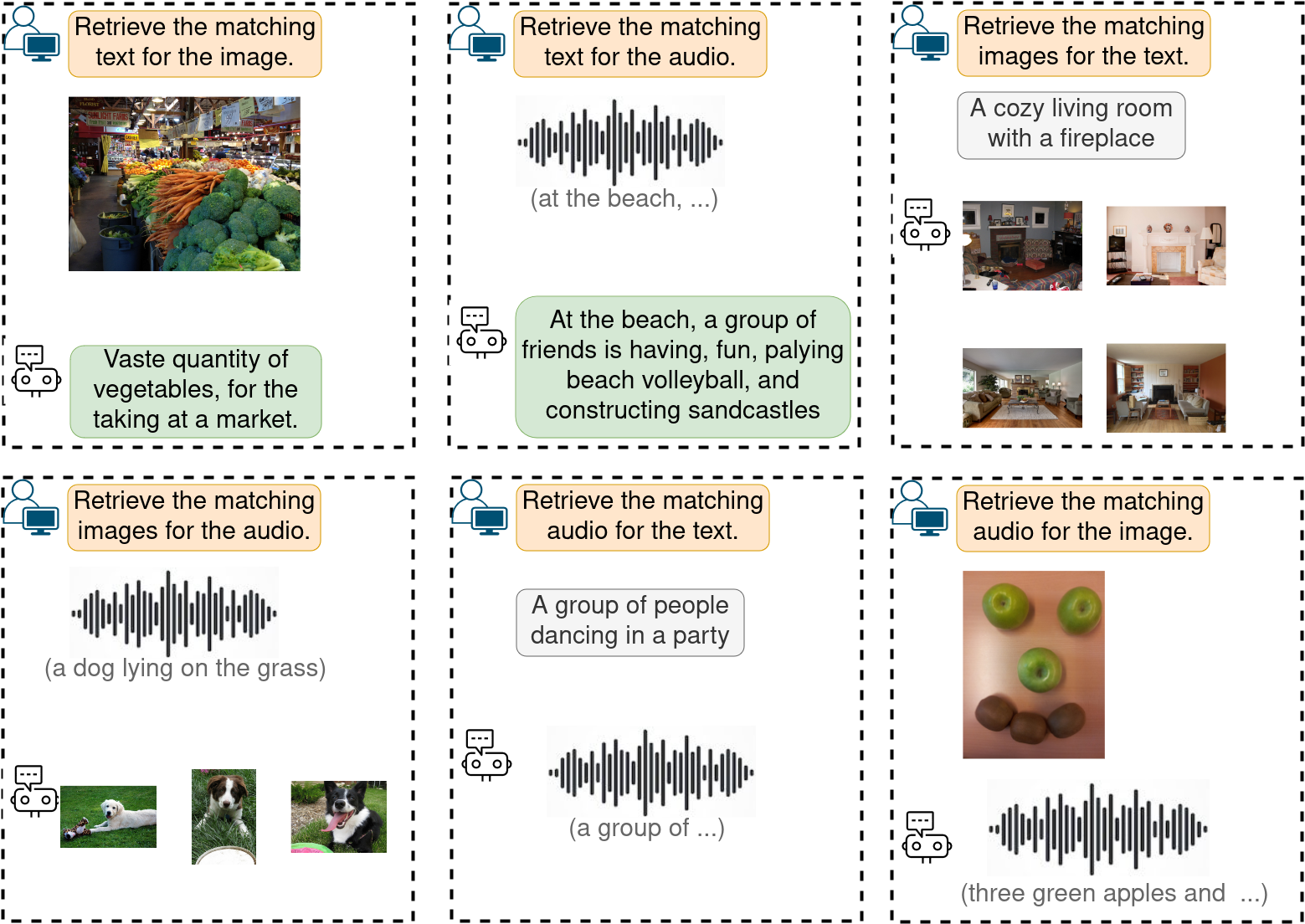}
      \caption{Qualitative results across three modalities. For each query, OneEncoder retrieves the most relevant data from the available dataset, showcasing the effectiveness of the alignment.}
      \label{fig:qualitative}
\end{figure*}
\indent Figure~\ref{fig:qualitative} presents qualitative results of OneEncoder across image, text, and audio modalities. In step 1 (see Algorithm~\ref{alg:one}), we demonstrate that OneEncoder effectively retrieves images using text queries and vice versa, indicating that the UP module can comprehend visual and textual content, resulting in relevant retrievals due to well-aligned latent space. In step 2 (see Algorithm~\ref{alg:two}), we show that image retrieval via audio inputs provides coherent data to the frozen UP, ensuring coherent alignment and making retrievals interesting. These illustrations, along with quantitative analysis, highlight OneEncoder's excellent performance in progressively aligning modalities using a lightweight framework. This framework enables coherent results with small aligned dataset training by utilizing frozen pretrained modality-specific encoders.

\section{OneEncoder ON VISUAL QUESTION ANSWERING}
In Section~\ref{sec:experiment}, we demonstrated that OneEncoder can be efficiently trained using a contrastive learning approach to align multiple modalities at a low computational cost. In this section, we introduce an alternative alignment method tailored for Visual Question Answering (VQA) tasks to further train OneEncoder. The goal is to illustrate the versatility of our proposed framework, showing its ability to be applied across various domains while utilizing different alignment strategies during training.\newline

VQA is a complex task that involves understanding both visual content and textual questions, requiring the model to align and reason across these modalities to generate accurate answers. By employing a specialized alignment mechanism for VQA, we aim to demonstrate OneEncoder's ability to handle cross-modal reasoning tasks beyond retrieval, further highlighting its adaptability across different types of multimodal learning challenges.
\begin{figure*}[!htbp]
    \centering
    \begin{subfigure}[b]{\textwidth}
        \centering
        \includegraphics[width=0.6\textwidth]{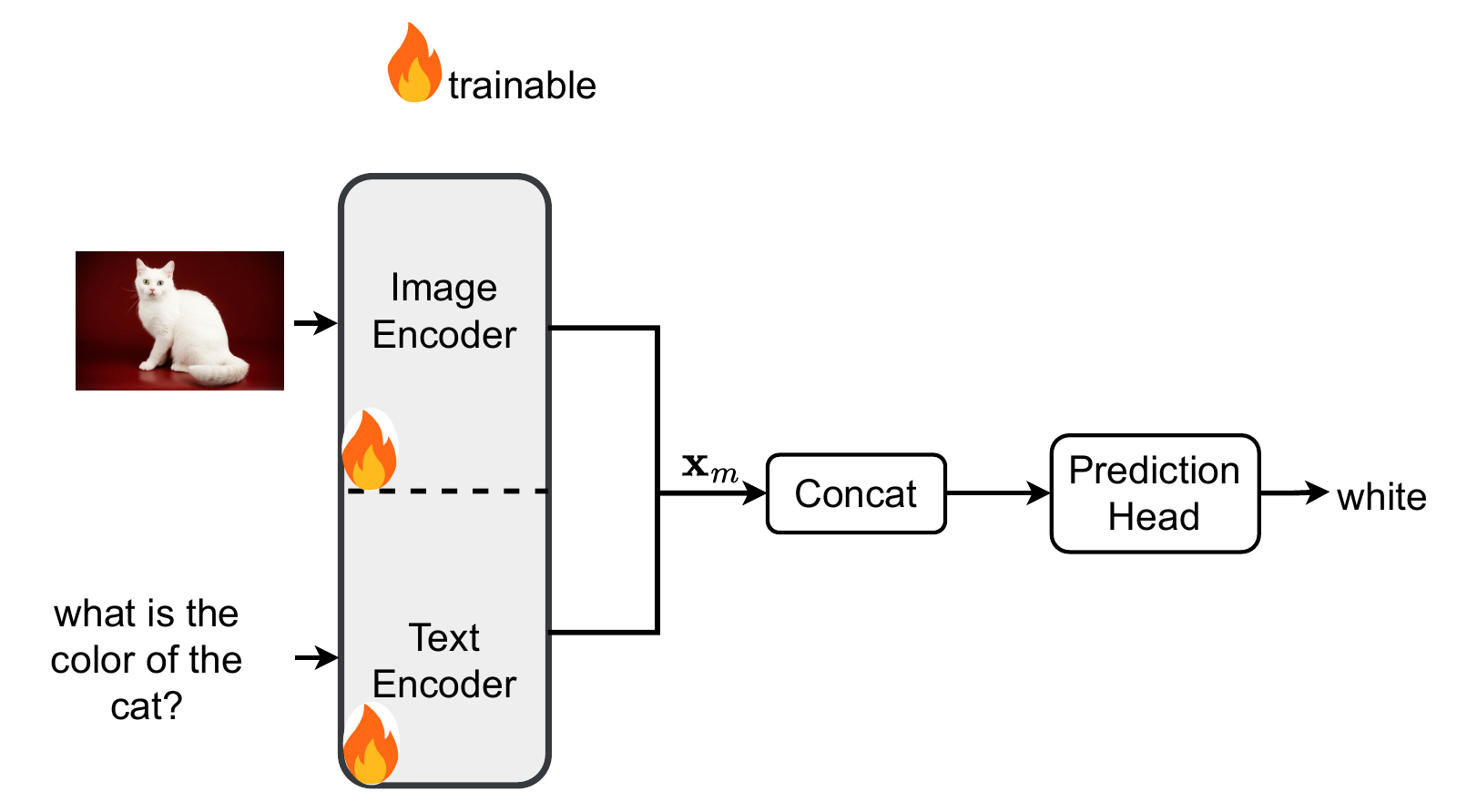} 
        \caption{\textbf{Baseline:} The parameters of both the Image Encoder and Text Encoder are trained.}
        \label{fig:vqa_baseline}
    \end{subfigure}
    \hfill
    \begin{subfigure}[b]{\textwidth}
        \centering
        \includegraphics[width=0.8\textwidth]{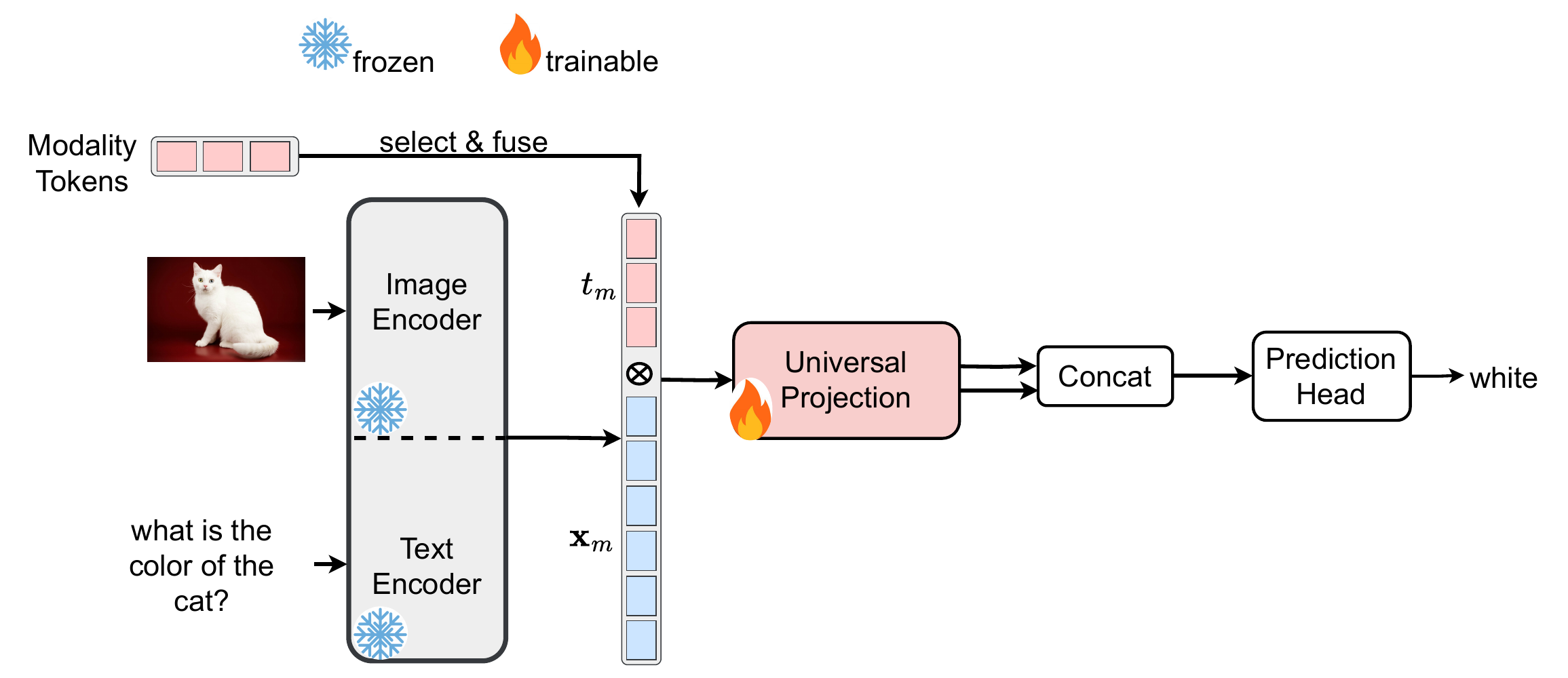} 
        \caption{\textbf{OneEncoder:} The parameters of both the Image Encoder and Text Encoder are frozen, with only the parameters of the Universal Projection (UP) module being trained.}
        \label{fig:vqa}
    \end{subfigure}
    \caption{\textbf{OneEncoder architecture for the Visual Question Answering (VQA) task.} The OneEncoder framework in~\ref{fig:vqa} trains only the UP module to align the textual answer with both the image and the textual question, unlike the baseline method in~\ref{fig:vqa_baseline}, which trains all specific encoders (image encoder and text encoder), making it more computationally expensive. Both approaches use a "Prediction Head" to generate textual answers.}
    \label{fig:oneencoder_vqa}
\end{figure*}
Figure~\ref{fig:oneencoder_vqa} presents a comparison between the classical VQA approach~\ref{fig:vqa_baseline} and the OneEncoder framework~\ref{fig:vqa}. As discussed in~\ref{sec:experiment}, OneEncoder trains only the UP module to align the textual answer with the image and question inputs, significantly reducing the number of parameters compared to the Baseline method~\ref{fig:vqa_baseline}, which requires training both the image and text encoders. Both methods utilize a "Prediction Head" module to generate the textual answer.\newline
\begin{figure}
    \centering
    \includegraphics[width=0.87\linewidth]{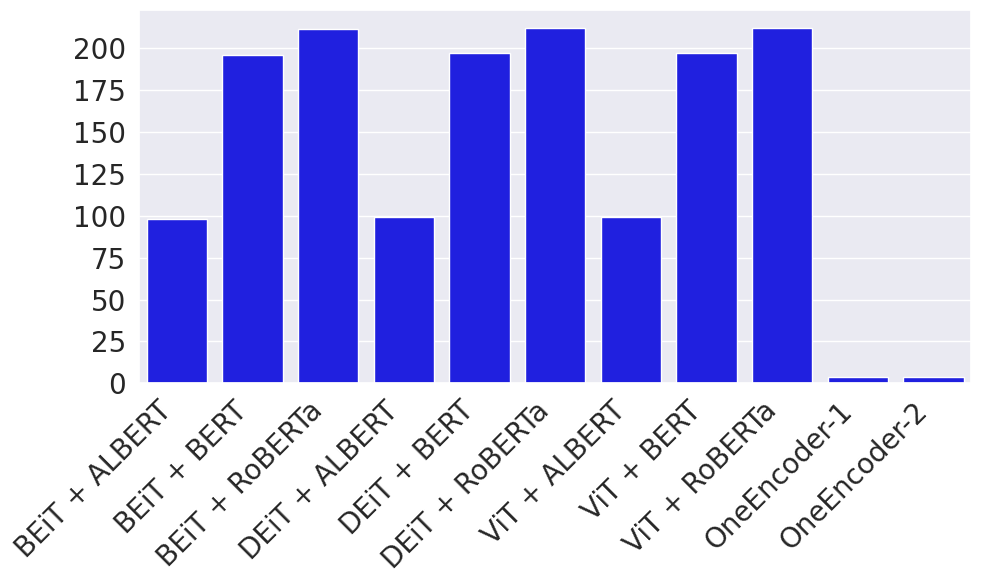}
    \caption{Comparison of Trainable Parameters (in millions) between Baseline Models and OneEncoder Variants (OneEncoder-1 and OneEncoder-2).}
    \label{fig:params}
\end{figure}

\begin{figure*}[!htbp]
    \centering
    \begin{subfigure}{\textwidth}
        \centering
        \begin{subfigure}{0.3\textwidth}
            \centering
            \includegraphics[width=\linewidth]{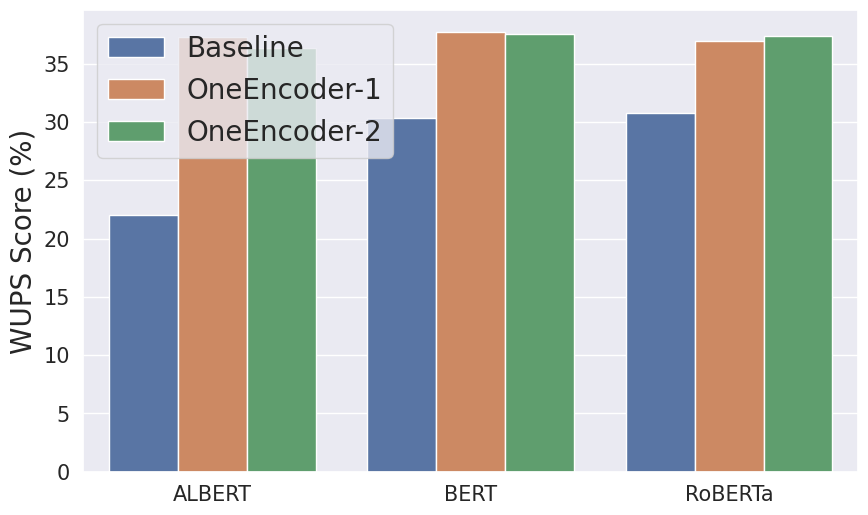} 
            \caption{BEiT}
        \end{subfigure}
        \begin{subfigure}{0.3\textwidth}
            \centering
            \includegraphics[width=\linewidth]{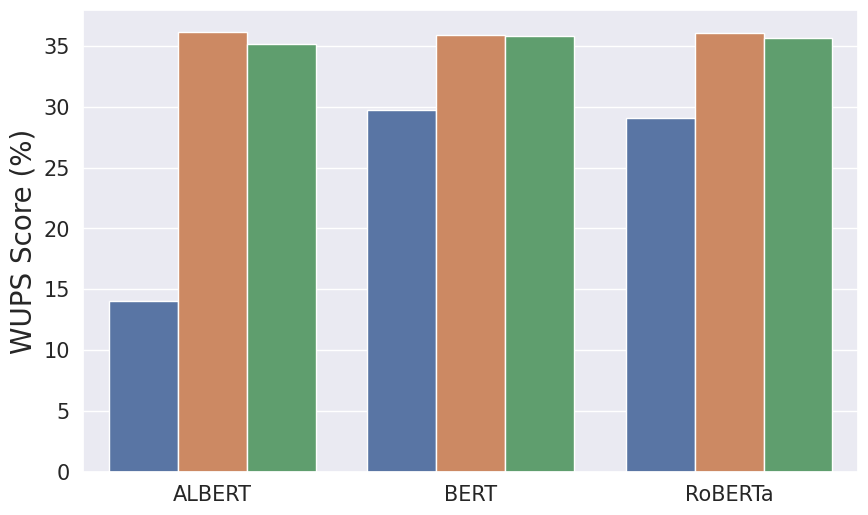} 
            \caption{DEiT}
        \end{subfigure}
        \begin{subfigure}{0.3\textwidth}
            \centering
            \includegraphics[width=\linewidth]{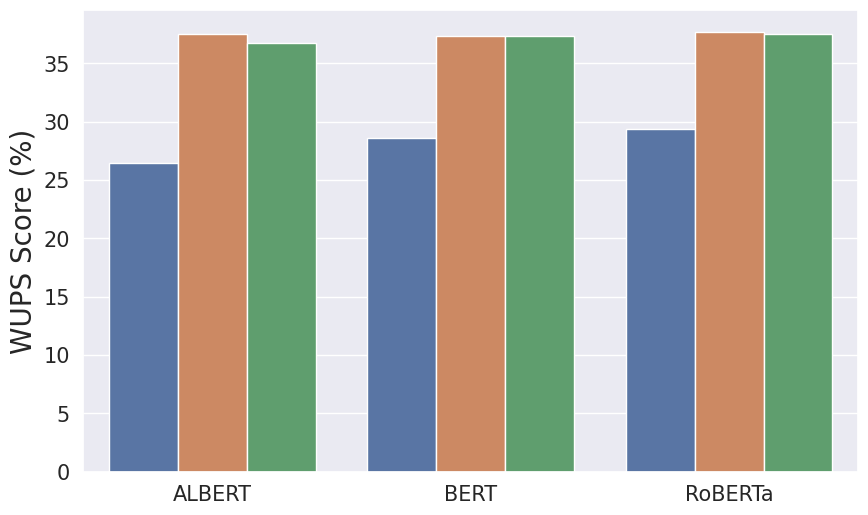} 
            \caption{ViT}
        \end{subfigure}
        \caption{WUPS score}
    \end{subfigure}
    \vspace{1em}     
    \begin{subfigure}{\textwidth}
        \centering
        \begin{subfigure}{0.3\textwidth}
            \centering
            \includegraphics[width=\linewidth]{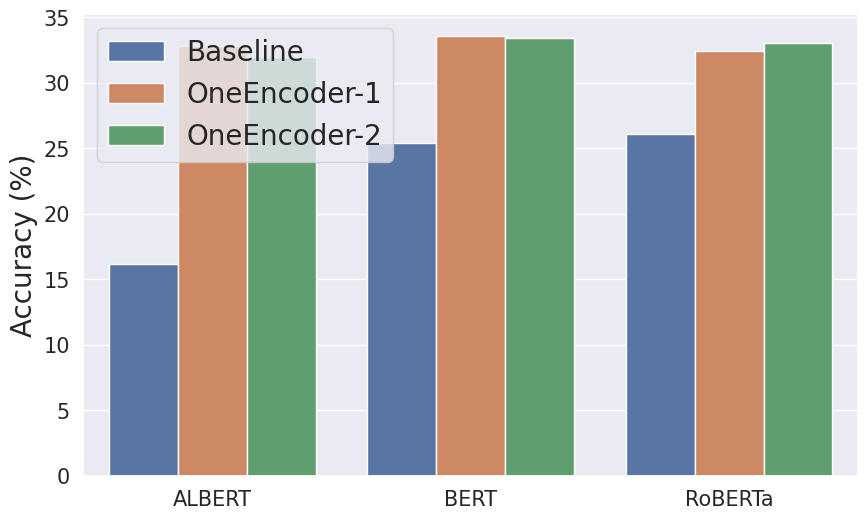} 
            \caption{BEiT}
        \end{subfigure}
        \begin{subfigure}{0.3\textwidth}
            \centering
            \includegraphics[width=\linewidth]{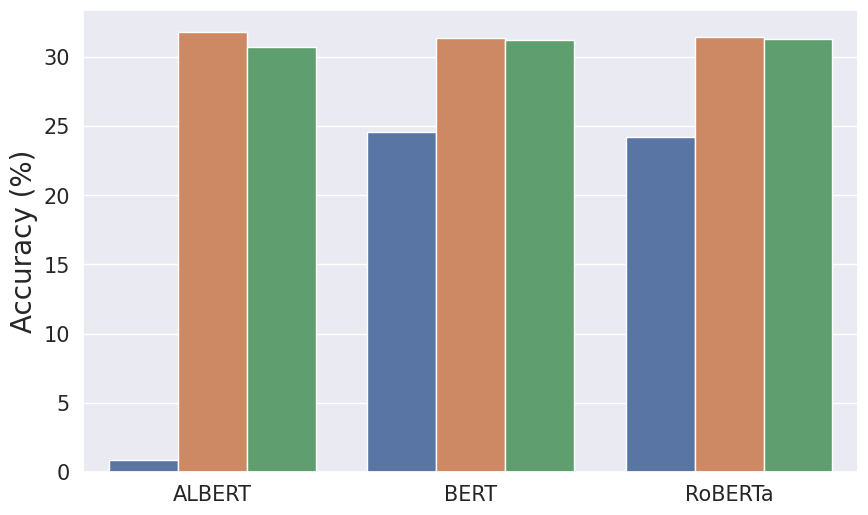} 
            \caption{DEiT}
        \end{subfigure}
        \begin{subfigure}{0.3\textwidth}
            \centering
            \includegraphics[width=\linewidth]{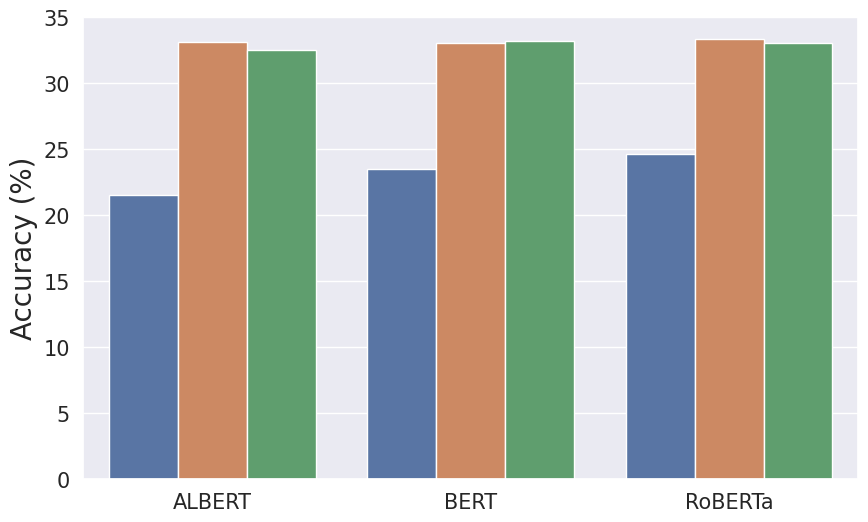} 
            \caption{ViT}
        \end{subfigure}
        \caption{Accuracy}
    \end{subfigure}
    \vspace{1em}
    \begin{subfigure}{\textwidth}
        \centering
        \begin{subfigure}{0.3\textwidth}
            \centering
            \includegraphics[width=\linewidth]{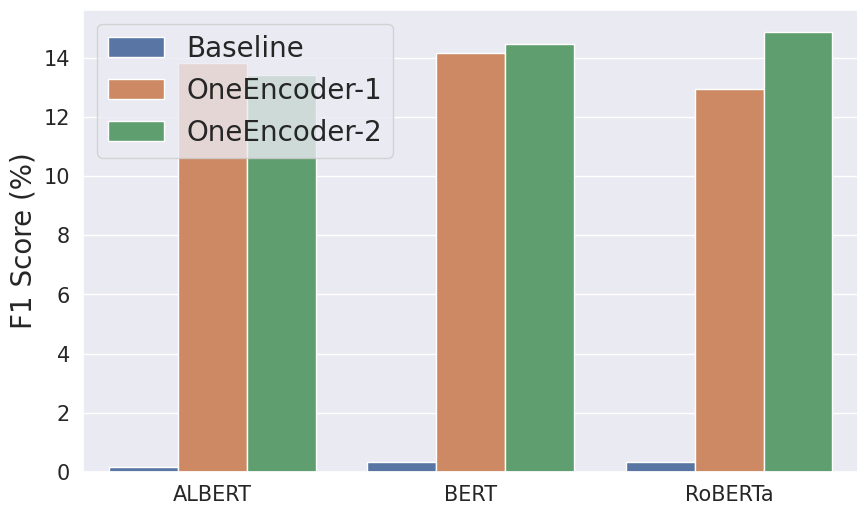} 
            \caption{BEiT}
        \end{subfigure}
        \begin{subfigure}{0.3\textwidth}
            \centering
            \includegraphics[width=\linewidth]{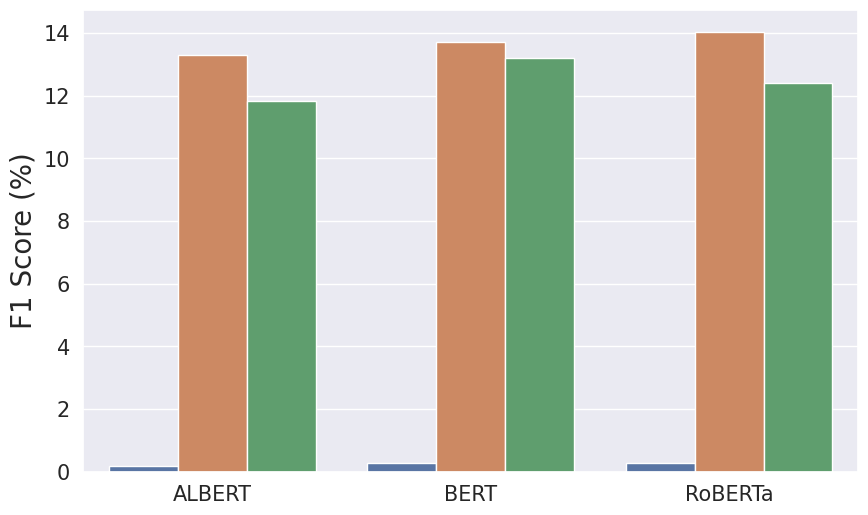} 
            \caption{DEiT}
        \end{subfigure}
        \begin{subfigure}{0.3\textwidth}
            \centering
            \includegraphics[width=\linewidth]{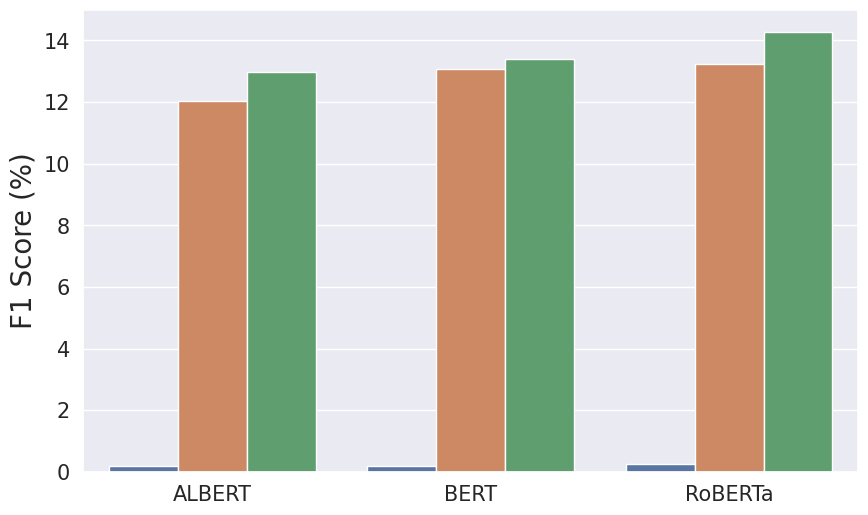}
            \caption{ViT}
        \end{subfigure}
        \caption{F1 score}
    \end{subfigure}
    \caption{Validation Performance of Baseline Models and OneEncoder Variants (OneEncoder-1, OneEncoder-2) on the DAQUAR dataset, evaluated using Wu-Palmer Similarity (WUPS), Accuracy, and F1 Score.}
    \label{fig:vqa_validation_metric}
\end{figure*}
\begin{figure*}
    \centering
    \includegraphics[width=0.87\linewidth]{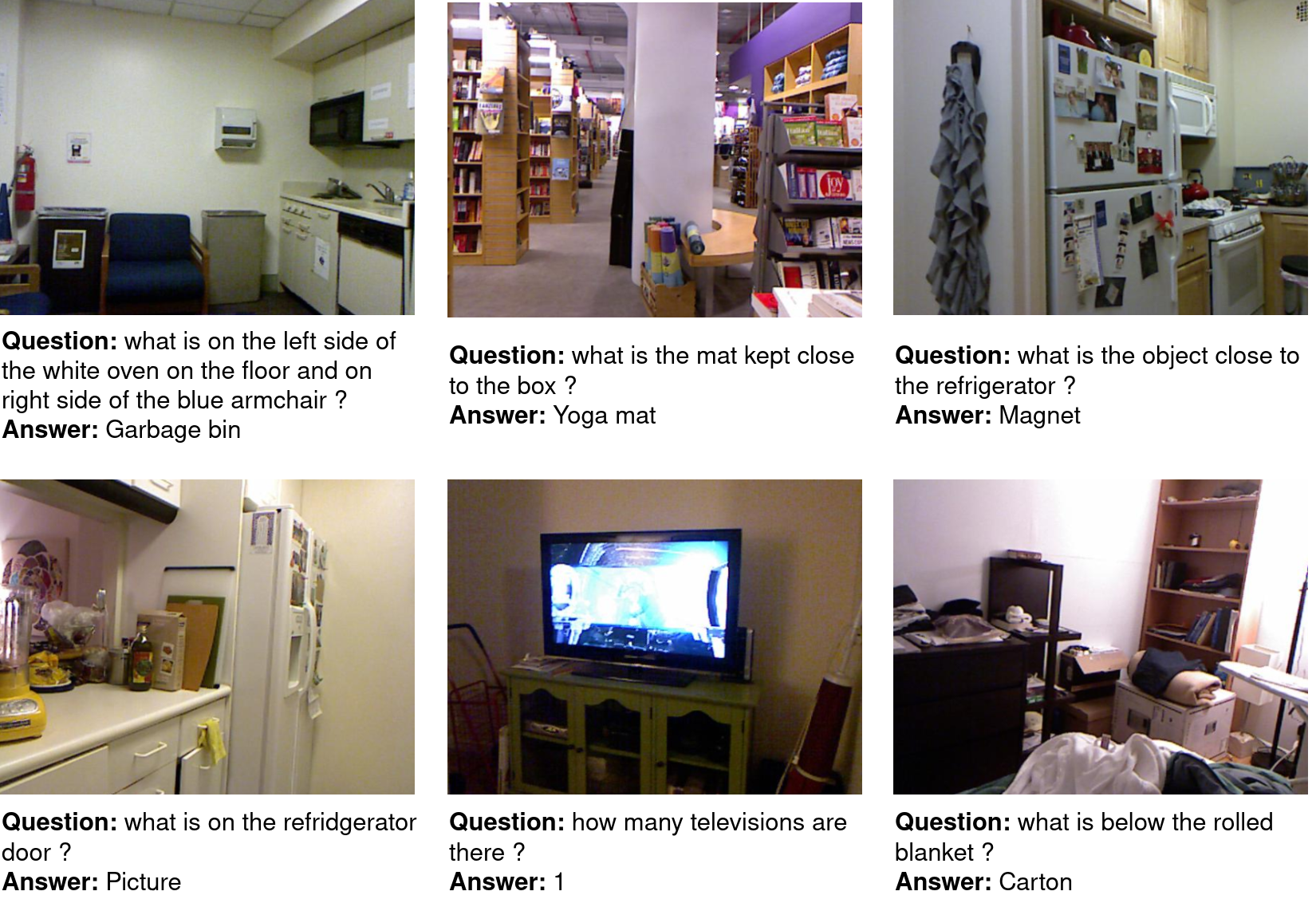}
    \caption{Example VQA Results Using the OneEncoder-1 Model.}
    \label{fig:vqa_example}
\end{figure*}
To train both the Baseline and OneEncoder frameworks, we utilize the DAQUAR (Dataset for Question Answering on Real-world images)~\cite{malinowski2014multi}. For modality-specific encoders, we employ BEiT-base~\cite{bao2021beit}, DEiT-base~\cite{touvron2021deit}, and ViT-base~\cite{orton2020vision} models as image encoders, each with 86M parameters, while ALBERT~\cite{lan2019albert} (60M parameters), BERT-base~\cite{devlin2019bert} (110M parameters), and RoBERTa-base~\cite{roberta} (125M parameters) serve as text encoders. We construct 9 VQA models for each method (Baseline and OneEncoder) by combining these encoder pairs: (BEiT, ALBERT), (BEiT, BERT), (BEiT, RoBERTa), (DEiT, ALBERT), (DEiT, BERT), (DEiT, RoBERTa), (ViT, ALBERT), (ViT, BERT), and (ViT, RoBERTa).\newline 

Since the DAQUAR dataset features simple vocabulary tokens as answers, we reformulate the task as a classification problem, using a linear layer as the "Prediction Head," where the output dimension matches the vocabulary size, and applying cross-entropy loss. Unlike the Baseline, which fine-tunes the entire pretrained modality-specific encoders, OneEncoder freezes these encoders and focuses solely on training the UP module. The goal of this application, using the smaller DAQUAR dataset, is to demonstrate that our framework can achieve strong performance with limited paired data, significantly reducing the number of parameters to optimize and shortening the training time required for convergence. We use four Transformer blocks with a total of 4M parameters for the UP module, and modality tokens of size $\mathbb{R}^{1 \times 768}$. All models are trained for 100 epochs without any data augmentation techniques.\newline

Figure~\ref{fig:params} provides a detailed comparison of the number of trainable parameters between Baseline models and OneEncoder variants. Specifically, OneEncoder-1 utilizes addition-based fusion, while OneEncoder-2 employs an attention-based fusion mechanism. Unlike the Baseline models, which train all parameters, the OneEncoder versions use Baseline models for feature extraction but keep them frozen during training.\newline

Figure~\ref{fig:vqa_validation_metric} demonstrates that the OneEncoder architecture (OneEncoder-1, OneEncoder-2) consistently outperforms baseline models across the three key metrics: Wu-Palmer Similarity (WUPS)~\cite{wu1994verbs}, Accuracy, and F1 Score. These results indicate that retraining specialized encoders may not be essential for achieving strong performance. By freezing the encoders and only training the UP on a small paired dataset, we can significantly reduce the number of parameters to optimize, minimize the need for large datasets, and shorten training times—all while yielding superior outcomes as illustrated in Figure~\ref{fig:vqa_example}.\newline

The VQA experiment further validates the findings in Section~\ref{sec:experiment}, focused on contrastive learning. OneEncoder, with its efficient and lightweight design, can be effectively integrated into any alignment-based approach, reducing parameter complexity, data requirements, and surpassing traditional methods that rely on retraining modality-specific encoders.

\section{CONCLUSION}
In this paper, we introduce a novel approach to multimodal representation learning: OneEncoder. Our method aims to reduce the training cost of multimodal systems by minimizing the number of parameters and the reliance on large datasets. OneEncoder leverages pretrained modality-specific encoders as fixed feature extractors, while only a lightweight, shared Universal Projection (UP) module is trained across all modalities. A modality token is incorporated before each representation to ensure consistent multimodal mapping within the UP.

For contrastive learning, OneEncoder aligns text and image embeddings within the same projection space, demonstrating superior performance on smaller datasets compared to more resource-intensive models like CLIP, which require separate training for each modality. To extend this framework to additional modalities such as audio and video, we propose a progressive alignment strategy. Here, the UP remains fixed while a compact Alignment Layer is introduced to map the output of modality-specific feature extractors into a UP-compatible space, providing a highly flexible and scalable solution without the need for large-scale retraining.

We further demonstrate OneEncoder's versatility through its application to visual question answering (VQA), where it achieves better performance and lower training costs compared to baseline models.

In conclusion, OneEncoder offers a scalable and parameter-efficient approach for multimodal representation learning, significantly reducing the need for extensive datasets while maintaining strong performance. Future work will explore its application to tasks like open-vocabulary object detection, where the model identifies relevant concepts based on rich semantic prompts before performing object detection.

\bibliographystyle{ieeetr}
\bibliography{bibliography}

\clearpage

\end{document}